\documentclass[runningheads]{llncs}
\usepackage{graphicx}
\usepackage{amsmath,amssymb} 
\usepackage{color}
\usepackage{times}
\usepackage{epsfig}
\usepackage{gensymb}
\usepackage{amsfonts}
\usepackage{booktabs}

\usepackage[width=122mm,left=12mm,paperwidth=146mm,height=193mm,top=12mm,paperheight=217mm]{geometry}

\newcommand{\KGthree}{\textcolor{black}}
\newcommand{\KG}{\textcolor{black}}
\newcommand{\KGtwo}{\textcolor{black}}
\newcommand{\cc}{\textcolor{black}}
\newcommand{\BX}{\textcolor{black}}
\newcommand{\bx}{\textcolor{black}}
\newcommand{\bo}{\textcolor{black}}
\newcommand{\boc}{\textcolor{black}}
\DeclareMathOperator*{\argmin}{argmin}

\begin{document}

\title{Snap Angle Prediction for 360$^{\circ}$ Panoramas} 

\titlerunning{Snap Angle Prediction for 360$^{\circ}$ Panorama}

\authorrunning{B. Xiong and K. Grauman}

\author{Bo Xiong\inst{1} \and Kristen Grauman\inst{2}}
\institute{University of Texas at Austin \\
\and  Facebook AI Research \\
\email{bxiong@cs.utexas.edu,grauman@fb.com}\footnote[1]{\emph{On leave from University of Texas at Austin (grauman@cs.utexas.edu).}}}

\maketitle

\begin{abstract}
360$^{\circ}$ panoramas are a rich medium, yet notoriously difficult to visualize in the 2D image plane.  We explore how intelligent rotations of a spherical image may enable content-aware projection with fewer perceptible distortions.  Whereas existing approaches assume the viewpoint is fixed, intuitively some viewing angles within the sphere preserve high-level objects better than others.  To discover the relationship between these optimal \emph{snap angles} and the spherical panorama's content, we develop a reinforcement learning approach for the cubemap projection model.  Implemented as a deep recurrent neural network, our method selects a sequence of rotation actions and receives reward for avoiding cube boundaries that overlap with important foreground objects. \boc{ We show our approach creates more visually pleasing panoramas while using 5x less computation than the baseline.}

\end{abstract}

\section{Introduction}\label{sec:introduction}

The recent emergence of inexpensive and lightweight 360$^{\circ}$ cameras enables exciting new ways to capture our visual surroundings.
Unlike traditional cameras that capture only a limited field of view, 360$^{\circ}$ cameras capture the entire visual world from their optical center.  
\boc{Advances in virtual reality  technology and promotion from social media platforms like YouTube and Facebook are further boosting the relevance of 360$^{\circ}$ data.}

However, viewing 360$^{\circ}$ content presents its own challenges.  
Currently three main directions are pursued: manual navigation, field-of-view (FOV) reduction, and content-based projection.
In manual navigation scenarios, a human viewer chooses which normal field-of-view subwindow to observe, e.g., via continuous head movements in a VR headset, or mouse clicks on a screen viewing interface.  
In contrast, FOV reduction methods generate normal FOV videos by learning to render the most interesting or capture-worthy portions of the viewing \boc{sphere}~\cite{su2016pano2vid,su2017making,hu2017deep,lai2017semantic}.  While these methods relieve the decision-making burden of manual navigation, they severely limit the information conveyed by discarding all unselected portions.  
Projection methods render a wide-angle view, or the entire sphere, onto a single plane (e.g., equirectangular or Mercator)~\cite{snyder1997flattening} or multiple planes~\cite{greene1986environment}.  While they avoid discarding content, any projection inevitably introduces distortions that can be unnatural for viewers.  Content-based projection methods can help reduce perceived distortions by prioritizing preservation of straight lines, conformality, or other low-level cues~\cite{sharpless2010pannini,kim2017automatic,li2015geodesic}, optionally using manual input to know what is worth preserving~\cite{carroll2009optimizing,tehrani2016correcting,carroll2010image,kopf2009locally,wang2015panorama}.

\begin{figure}[t]
\centering
\renewcommand{\tabcolsep}{0pt}
\includegraphics[width=1\columnwidth]{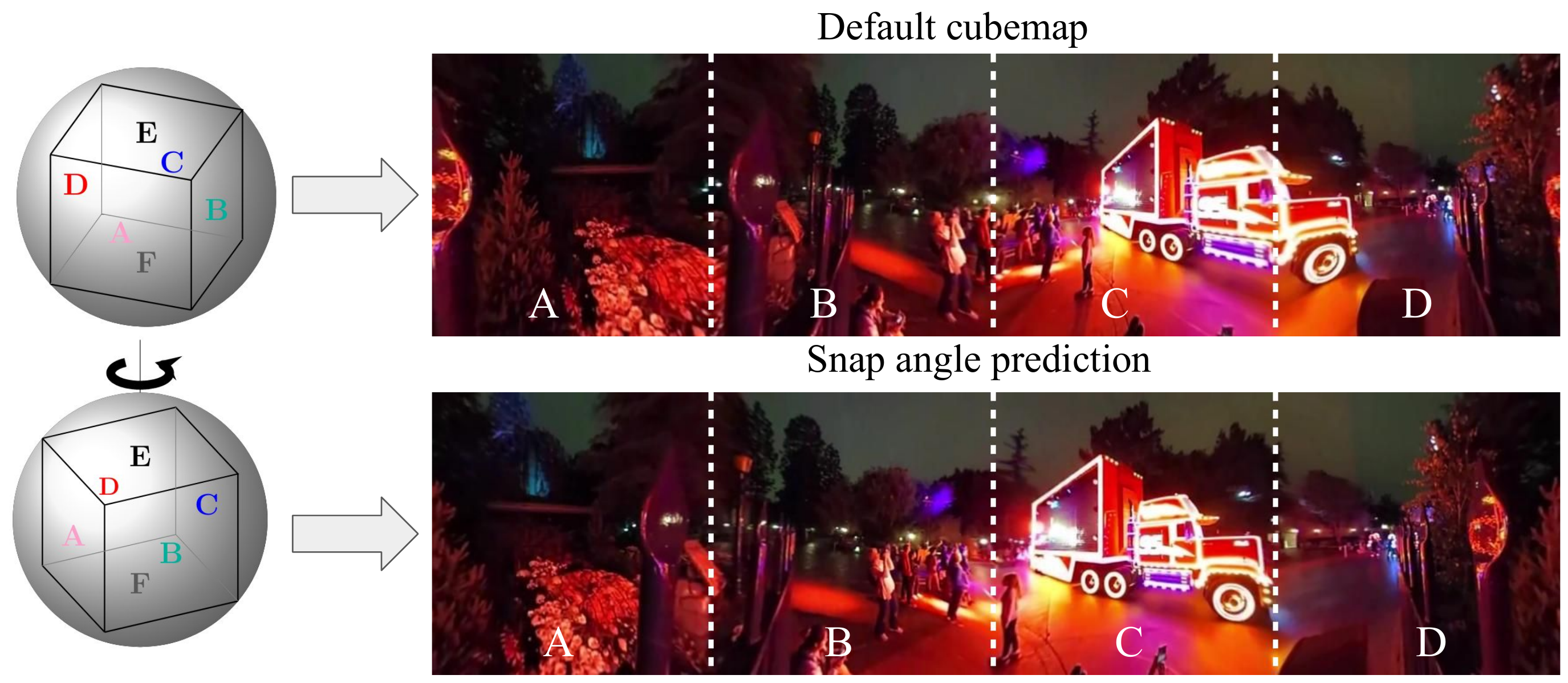}
\caption{Comparison of a cubemap before and after snap angle prediction (dotted lines separate each face). 
Unlike prior work that assumes a fixed angle for projection, we propose to predict the cube rotation that will best preserve foreground objects in the output.  For example, here our method better preserves
the truck (third picture C in the second row).  We show four (front, right, left, and back) out of the six faces for visualization purposes. Best viewed in color or pdf.}
\label{fig:intro}
\vspace{-13pt}
\end{figure}

However, all prior \BX{automatic} content-based projection methods implicitly assume that the \emph{viewpoint} of the input 360$^{\circ}$ image is fixed.  That is, the spherical image is processed in some default coordinate system, e.g., \KG{as the equirectangular projection provided by the camera manufacturer.} This assumption limits the quality of the output image.  Independent of the content-aware projection eventually used, a fixed viewpoint means some \emph{arbitrary portions of the original sphere will be relegated to places where distortions are greatest}---or at least where they will require most attention by the content-aware algorithm to ``undo''.  

We propose to eliminate the fixed viewpoint assumption.  Our key insight is that an intelligently chosen viewing angle can immediately lessen distortions, even when followed by a conventional projection approach.  In particular, we consider the widely used \boc{cubemap} projection~\cite{greene1986environment,fb2015cubemap,google2017eac}.  A cubemap visualizes the entire sphere by first mapping the sphere to a cube with rectilinear projection (where each face captures a 90$^{\circ}$ FOV) and then unfolding the faces of the cube.  Often, an important object can be projected across two cube faces, destroying object integrity.  In addition, rectilinear projection distorts content near cube face boundaries more.  See Figure~\ref{fig:intro}, top.  However, intuitively, some viewing angles---some cube orientations---are less damaging than others.

We introduce an approach to automatically predict \emph{snap angles}: the rotation of the cube that will yield a set of cube faces that, among all possible rotations, most look like nicely composed human-taken photos originating from the given 360$^{\circ}$ panoramic image.   While what comprises a ``well-composed photo'' is itself the subject of active research~\cite{kong2016aesthetics,isola2011makes,xiong2014detecting,gygli2013interestingness,ICCV15_Khosla}, we concentrate on a high-level measure of good composition, where the goal is to consolidate each (automatically detected) foreground object within the bounds of one cubemap face.   See Figure~\ref{fig:intro}, bottom.

Accordingly, we formalize our snap angle objective in terms of minimizing the spatial mass of foreground objects near cube edges.  We develop a reinforcement learning (RL) approach to infer the optimal snap angle given a 360$^{\circ}$ panorama.  We implement the approach with a deep recurrent neural network that is trained end-to-end.  The sequence of rotation ``actions'' chosen by our RL network can be seen as a \boc{\emph{learned}} coarse-to-fine adjustment of the camera viewpoint, in the same spirit as how people refine their camera's orientation just before snapping a photo.

We validate our approach on a variety of 360${^\circ}$ panorama \KGthree{images}.  Compared to several informative baselines, we demonstrate that  1) snap angles better preserve important objects, 2) our RL solution efficiently pinpoints the best snap angle, 3) cubemaps unwrapped after snap angle rotation suffer less perceptual distortion than the status quo cubemap, and 4) snap angles even have potential to impact recognition applications, by orienting 360$^{\circ}$ data in ways that better match the statistics of normal FOV photos used for today's pretrained recognition networks.

\section{Related Work} \label{sec:related}

\paragraph{Spherical image projection}

Spherical image projection models project either a limited FOV~\cite{sharpless2010pannini,chang2013rectangling} or the entire panorama~\cite{snyder1997flattening,zelnik2005squaring,greene1986environment}.  The former group includes rectilinear and Pannini~\cite{sharpless2010pannini} projection; the latter includes equirectangular, stereographic, and Mercator projections (see~\cite{snyder1997flattening} for a review).  \BX{Rectilinear and Pannini} 
prioritize preservation of lines in various ways, \KG{but always independent of the specific input image}. Since any projection of the full sphere must incur distortion, multi-view projections can be perceptually stronger than a single global projection~\cite{zelnik2005squaring}.  Cubemap~\cite{greene1986environment}, the subject of our snap angle approach, is a multi-view projection method; as discussed above, current approaches simply consider a cubemap in its default orientation.

\vspace*{-0.1in}
\paragraph{Content-aware projection}
Built on spherical projection methods, content-based projections make \KG{image-specific choices} to reduce distortion.
Recent work~\cite{kim2017automatic} optimizes the parameters in the Pannini projection~\cite{sharpless2010pannini} to preserve regions with greater low-level saliency and straight lines.  Interactive methods~\cite{carroll2009optimizing,tehrani2016correcting,carroll2010image,kopf2009locally} require a user to outline regions of interest that should be preserved
\BX{or require input from a user to determine projection orientation~\cite{wang2015panorama}.}
Our approach is content-based \KGthree{and fully automatic}.  Whereas prior automatic methods assume a fixed viewpoint for projection, we propose to actively predict snap angles for rendering.  
Thus, our idea is orthogonal to 360$^{\circ}$ content-aware projection.  Advances in the projection method could be applied in concert with our algorithm, \KG{e.g., as post-processing to enhance the rotated faces further.}  For example, when generating cubemaps, one could replace rectilinear projection with others~\cite{sharpless2010pannini,kim2017automatic,carroll2009optimizing} and keep the rest of our \BX{learning} framework unchanged.
Furthermore, the proposed snap angles respect high-level image content---detected foreground objects---as opposed to typical lower-level cues like line straightness~\cite{carroll2010image,carroll2009optimizing} or low-level saliency metrics~\cite{kim2017automatic}.

\vspace*{-0.1in}
\paragraph{Viewing wide-angle panoramas}
Since viewing 360$^{\circ}$ and wide-angle data is non-trivial, there are vision-based efforts to facilitate.  The system of~\cite{kopf2007capturing} helps efficient exploration of gigapixel panoramas.  
More recently, several systems automatically extract normal FOV videos from 360$^{\circ}$ video, ``piloting'' a virtual camera by selecting the viewing angle and/or zoom level most likely to interest a human viewer~\cite{su2016pano2vid,su2017making,hu2017deep,lai2017semantic}.

\vspace*{-0.1in}
\paragraph{\KG{Recurrent networks for attention}}

\KG{Though treating very different problems than ours, multiple recent methods incorporate deep recurrent neural networks (RNN) to make sequential decisions about where to focus attention.}  The influential work of~\cite{mnih2014recurrent}
  learns a policy for visual attention in image classification.  \boc{Active perception systems use RNNs and/or reinforcement learning to select places to look in a novel image~\cite{caicedo2015active,mathe2016reinforcement}, environment~\cite{jayaraman2016look,jayaraman2017learning,jayaraman2018end}, or video~\cite{yeung2016end,alwassel2017action,singh2016multi,su2016leaving} to detect certain objects or activities efficiently.} Broadly construed, we share the general goal of efficiently converging on a desired target ``view'', but our problem domain is entirely different.

\vspace*{-0.1in}
\section{Approach} \label{sec:approach}
\vspace*{-0.05in}
We first formalize  snap angle prediction as an optimization problem (Sec.~\ref{sec:snap_angle}). Then present our learning framework and network architecture for snap angle prediction (Sec.~\ref{sec:model}).

We concentrate on the cubemap projection~\cite{greene1986environment}.
Recall that a cubemap maps the sphere to a cube with rectilinear projection (where each face captures a 90$^{\circ}$ FOV) and then unfolds the six faces of the cube.  The unwrapped cube can be visualized as an unfolded box, with the lateral strip of four faces being spatially contiguous in the scene (see Fig.~\ref{fig:intro}, bottom).  We explore our idea with cubemaps for a couple reasons.  First, a cubemap covers the entire 360$^{\circ}$ content and does not discard any information.  Secondly, each cube face is very similar to a conventional 
\BX{FOV}, and therefore relatively easy for a human to view \KGthree{and/or} edit.

\vspace*{-0.05in}
\subsection{Problem Formulation}\label{sec:snap_angle}

We first formalize snap angle prediction as an optimization problem.  
Let $P(I,\theta)$ denote a projection function that takes a panorama image $I$ and a projection angle $\theta$ as input and outputs a cubemap \KG{after rotating the sphere (or equivalently the cube) by $\theta$}. Let function $F$ be an objective function that takes a cubemap as input and outputs a score to measure the quality of the cubemap. Given a novel panorama image $I$, our goal is to minimize $F$ by predicting the snap angle $\theta^\ast$:
\begin{equation}
\begin{matrix}
\displaystyle \KG{\theta^\ast}  = \argmin_\theta & F(P(I,\theta)).  \\
\end{matrix}
\end{equation}
The projection function $P$ first transforms the coordinates of each point in the panorama based on the snap angle $\theta$ and then produces a cubemap in the standard manner.

Views from a horizontal camera position (elevation 0$^{\circ}$) are more informative than others due to human recording bias. The bottom and top cube faces often align with the sky (above) and  ground (below); ``stuff'' regions like sky, ceiling, and floor are thus common in these faces and foreground objects are minimal. \KGthree{Therefore, rotations in azimuth tend to have greater influence on the disruption caused by cubemap edges.}  Hence, without loss of generality, we focus on snap angles in azimuth only, and jointly optimize the front/left/right/back faces of the cube.

The coordinates for each point in a panorama can be represented by a pair of 
\BX{latitude and longitude}
$(\lambda,\varphi)$.  Let $L$ denote a coordinate transformation function that takes the snap angle $\theta$ and a pair of coordinates as input.  We define the coordinate transformation function $L$ as:
\begin{equation}
\begin{matrix}
L((\lambda,\varphi),\theta)=(\lambda,\varphi-\theta).
  \\
\end{matrix}
\label{eq:L}
\end{equation}
Note when the snap angle is 90$^{\circ}$, the orientation of the cube is the same as the default cube except the order of front, back, right, and left is changed. We therefore restrict $\theta \in [0,\pi /2]$. We discretize the space of candidate angles for $\theta$ into a uniform \KG{$N=20$} azimuths grid, \BX{which we found offers fine enough camera control.}

\begin{figure*}[t]
\centering
\renewcommand{\tabcolsep}{0pt}
\includegraphics[width=0.95\columnwidth]{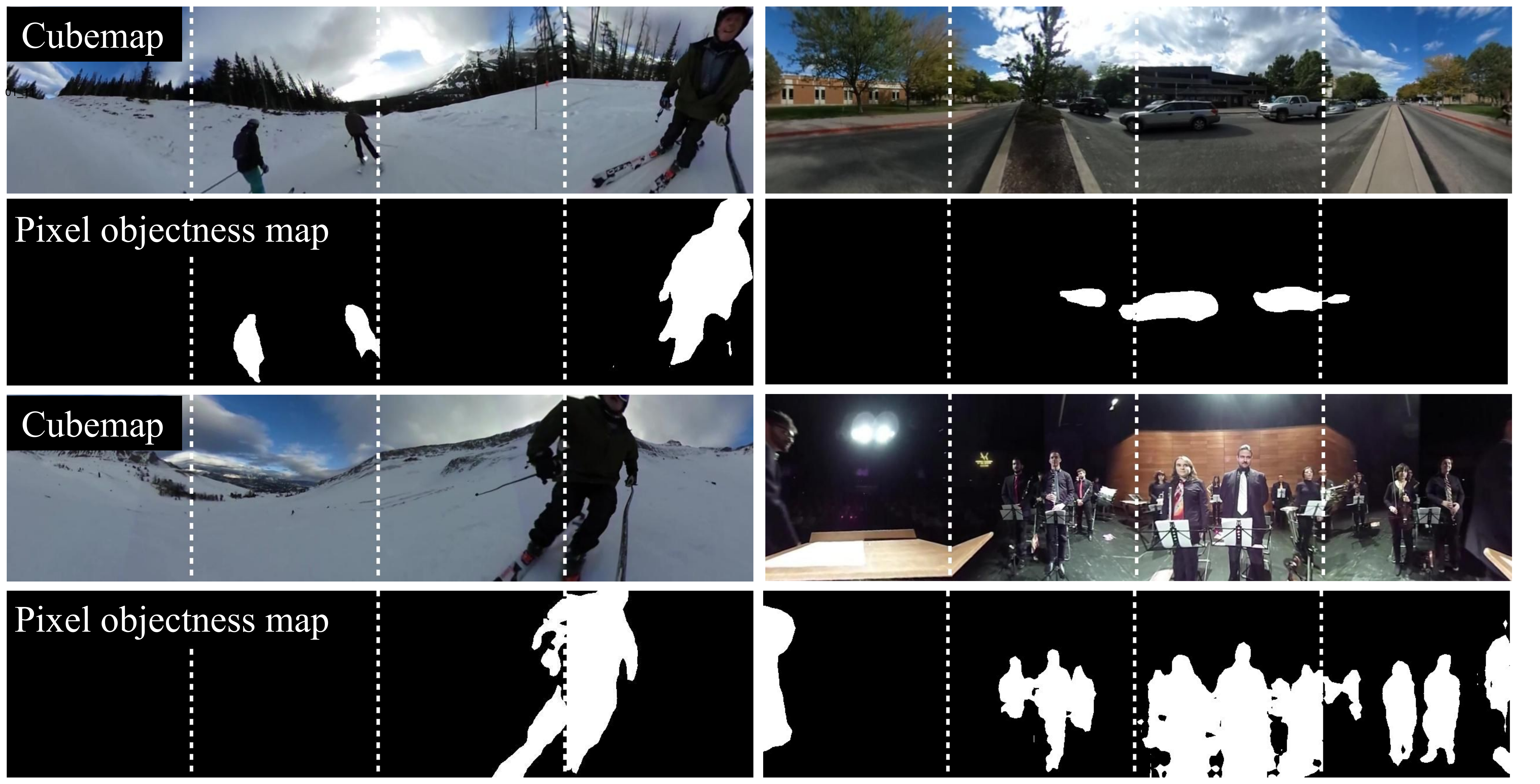}
\caption{Pixel objectness~\cite{jain2017pixel} foreground map examples. White pixels in the pixel objectness map indicate foreground.  \KGthree{Our approach learns to find cubemap orientations where the foreground objects are not disrupted by cube edges, i.e., each object falls largely within one face.}}
\label{fig:pixel}
\vspace{-0.2in}
\end{figure*}

We next discuss our choice of the objective function $F$.  
A cubemap in its default orientation has two disadvantages: 1) It does not guarantee to project each important object onto the same cube face; 2) Due to the nature of the perspective projection, objects projected onto cube boundaries will be distorted more than objects in the center.  Motivated by these shortcomings, our goal is to produce cubemaps that \emph{place each important object in a single face} and avoid placing objects at the cube boundaries/edges. 

In particular, we propose to minimize the area of foreground objects near or on cube boundaries. 
Supposing each pixel in a cube face is \boc{automatically} labeled as either object or background, our objective $F$ measures \emph{the fraction of pixels that are labeled as \BX{foreground} near cube boundaries.}  A pixel is near cube boundaries if it is less than \KG{$A$}\% of the cube length away from the left, right, or top boundary. We do not penalize objects near the bottom boundary since it is common to place objects near the bottom boundary in photography (e.g., potraits).

To infer which pixels belong to the foreground, we use ``pixel objectness''~\cite{jain2017pixel}.  Pixel objectness is a CNN-based foreground estimation approach that returns pixel-wise estimates for all foreground object(s) in the scene, no matter their category.  While other foreground methods are feasible (e.g.,~\cite{zitnick2014edge,carreira2012cpmc,jiang2013,pinheiro2015learning,liu2011learning}), we choose pixel objectness due to its accuracy in detecting foreground objects of any category, as well as its ability to produce a single \BX{pixel-wise foreground map which can contain multiple objects.}
\KGthree{Figure~\ref{fig:pixel} shows example pixel objectness foreground maps on cube faces.}
\KGthree{We apply pixel objectness to a given projected cubemap to obtain its pixel objectness score.}
In conjunction, other measurements for photo quality, such as interestingness~\cite{gygli2013interestingness}, memorability~\cite{isola2011makes}, or aesthetics~\cite{dhar2011high}, could be employed within $F$.

\subsection{Learning to Predict Snap Angles}\label{sec:model}

On the one hand, a direct regression solution would attempt to infer $\theta^\ast$ directly from $I$.  However, this is problematic because good snap angles can be multi-modal, \boc{i.e.,} available at multiple directions in the sphere, and thus poorly suited for regression.  
On the other hand, a brute force solution would require projecting the panorama to a cubemap and then evaluating $F$ for every possible projection angle $\theta$, which is costly.

We instead address snap angle prediction with reinforcement learning.  The task is a time-budgeted sequential decision process---an iterative adjustment of the (virtual) camera rotation that homes in on the least distorting viewpoint for cubemap projection.  Actions are cube rotations and rewards are improvements to the pixel objectness score $F$.  Loosely speaking, this is reminiscent of how people take photos with a coarse-to-fine refinement towards the desired composition.  \KGthree{However, unlike a naive coarse-to-fine search, our approach learns to trigger different search strategies depending on what is observed, as we will demonstrate in results.}

Specifically, let $T$ represent the budget given to our system, indicating the number of rotations it may attempt.  We maintain a history of the model's previous predictions. At each time step $t$, our framework takes a relative snap prediction $s_t$ (for example, $s_{t}$ could  signal to update the azimuth by 45$^{\circ}$) and updates its previous snap angle $\theta_{t}=\theta_{t-1}+s_t$. Then, \KG{based on its current observation}, our system makes a prediction $p_t$, which is used to update the snap angle in the next time step. That is, we have $s_{t+1}=p_t$. Finally, we choose the snap angle 
with the lowest pixel objectness objective score from the history as our final prediction $\hat{\theta}$:

\begin{equation}
\hat{\theta}=\argmin_{\theta_t=\theta_1,...,\KG{\theta_T}} F(\KG{P}(I,\theta_t)).
\end{equation}

\KGthree{To further improve efficiency, one could compute pixel objectness \emph{once} on a cylindrical panorama rather than recompute it for every cubemap rotation, and then proceed with the iterative rotation predictions above unchanged.  However, learned foreground detectors~\cite{jain2017pixel,jiang2013,carreira2012cpmc,pinheiro2015learning,liu2011learning} are trained on Web images in rectilinear projection, and so their accuracy can degrade with different distortions.  Thus we simply recompute the foreground for each cubemap reprojection.  See Sec.~\ref{sec:pred_angle} for run-times.}

\begin{figure*}[t]
\centering
\renewcommand{\tabcolsep}{0pt}
\includegraphics[width=1\columnwidth]{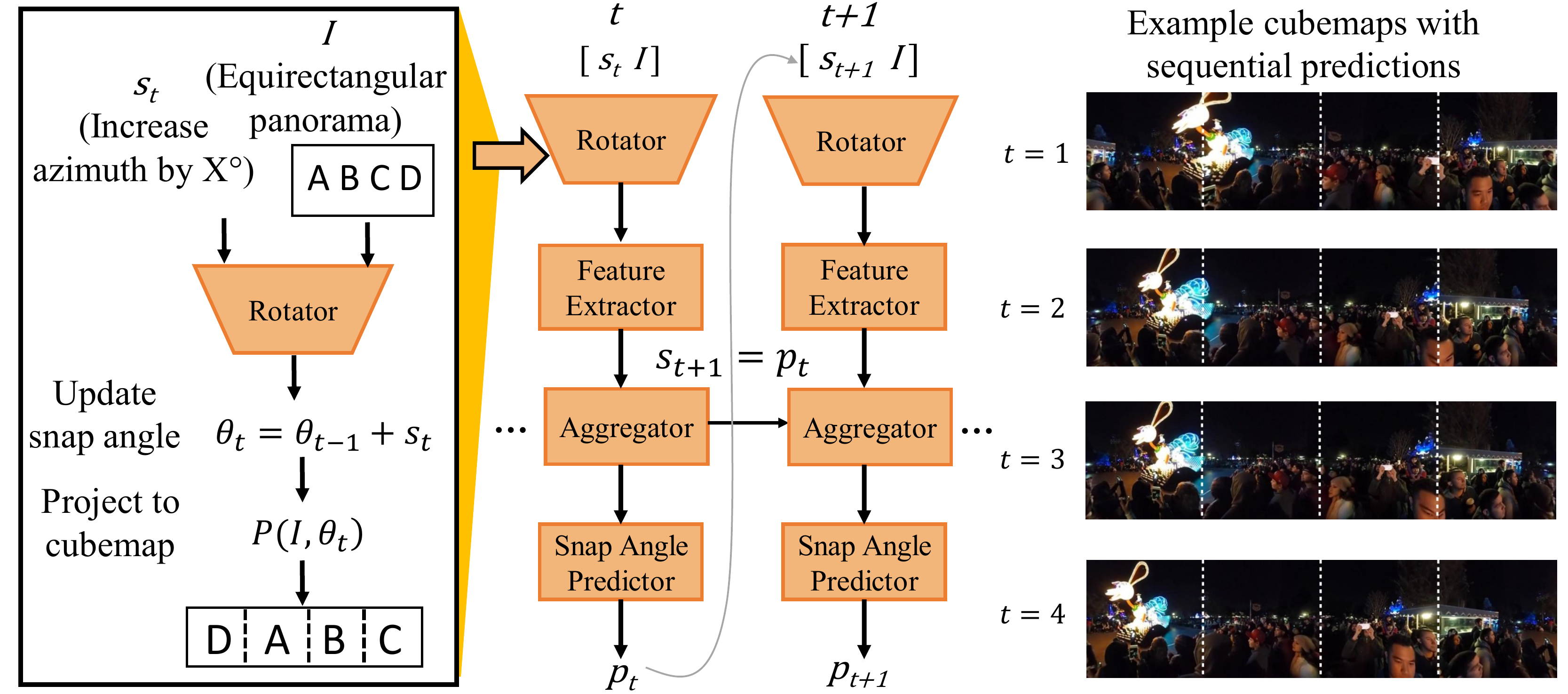}
\caption{We show the rotator (left), our model (middle), and a series of cubemaps produced by our sequential predictions (right). Our method iteratively refines the best snap angle, targeting a given budget of allowed computation.
}
\label{fig:net}
\vspace{-0.15in}
\end{figure*}

\vspace*{-0.1in}
\paragraph{Network}  We implement our reinforcement learning task \boc{with} deep recurrent and convolutional neural networks.    
Our framework consists of four modules: a \textit{rotator}, a \textit{feature extractor}, an \textit{aggregator}, and a \textit{snap angle predictor}. At each time step, it processes the data and produces a cubemap (\textit{rotator}), extracts learned features (\textit{feature extractor}), integrates information over time (\textit{aggregator}), and predicts the next snap angle (\textit{snap angle predictor}).

At each time step $t$, the \textit{rotator} takes as input a panorama $I$ in equirectangular projection and a relative snap angle prediction $s_{t}=p_{t-1}$, which is the prediction from the previous time step. The \textit{rotator} updates its current snap angle prediction with $\theta_t=\theta_{t-1}+s_{t}$. We set $\theta_1=0$ initially.
Then the \KGtwo{\textit{rotator}} applies the projection function $P$ to $I$ based on $\theta_t$ with Eq~\ref{eq:L} to produce a cubemap. 
Since our objective is to minimize the total amount of foreground straddling cube face boundaries, it is more efficient for our model to learn directly from the pixel objectness map than from raw pixels. 
Therefore, we apply pixel objectness~\cite{jain2017pixel} to each of the four lateral cube faces to obtain a binary objectness map per face. 
The rotator has the form:
$\mathbb{I} ^{W \times H \times 3}\times \Theta \rightarrow \mathbb{B}^{W_c \times W_c \times 4}$,
%
where $W$ and $H$ are the width and height of the input panorama in equirectangular projection and $W_c$ denotes the side length of a cube face. The \textit{rotator} does not have any learnable parameters since it is used to preprocess the input data.

At each time step $t$, the \textit{feature extractor} then applies a sequence of convolutions to the output of the \textit{rotator} to produce a feature vector $f_t$, which is then fed into the \textit{aggregator} to produce an aggregate feature vector $a_t=A(f_1,...,f_t)$ over time. Our \textit{aggregator} is a recurrent neural network (RNN), which also maintains its own hidden state.

Finally, the \textit{snap angle predictor} takes the aggregate feature vector as input, and produces a relative snap angle prediction $p_{t}$. In the next time step $t+1$, the relative snap angle prediction is fed into the \textit{rotator} to produce a new cubemap. The \textit{snap angle predictor} contains two fully connected  layers, each followed by a ReLU, and then the output is fed into a softmax function \KG{for the $N$ azimuth candidates}. 
The $N$ candidates here are relative, and range from decreasing azimuth by $\frac{N}{2}$ to increasing azimuth by $\frac{N}{2}$.  
The \emph{snap angle predictor} first produces a multinomial probability density function $\pi(p_t)$ over all candidate relative snap angles, then it samples one snap angle prediction proportional to the probability density function.  See Figure~\ref{fig:net} for an overview of the network, and \BX{Supp.} 
 for all architecture details.

\vspace*{-0.1in}
\paragraph{Training}
The parameters of our model consist of parameters of the \textit{feature extractor}, \textit{aggregator}, and \textit{snap angle predictor}: $w=\{w_f, w_a, w_p\}$.  We learn them to maximize the total reward (defined below) our model can expect when predicting snap angles. The \textit{snap angle predictor} contains stochastic units and therefore cannot be trained with the standard backpropagation method.  We therefore use REINFORCE~\cite{williams1992simple}.  Let $\pi(p_t|I, w)$ denote the parameterized policy, which is a pdf over all possible snap angle predictions. REINFORCE iteratively increases weights in the pdf $\pi(p_t|I, w)$ on those snap angles that have received higher rewards. Formally, given a batch of training data $\{I_i:i=1,\dots,M\}$, we can approximate the gradient as \boc{follows}:
\begin{equation}
\sum_{i=1}^{\KG{M}} \sum_{t=1}^{T} \nabla_w \log \pi(p^i_t| I_{i}, w)R^i_t
\end{equation}
\KG{where $R^i_t$ denotes the reward at time $t$ for instance $i$.}

\vspace*{-0.1in}
\paragraph{Reward}

At each time step $t$, we compute the objective.  Let $\hat\theta_t=\argmin_{\theta=\theta_1,\dots \theta_t}F(P(I, \theta))$ denote the snap angle with the lowest pixel objectness until time step $t$. Let $O_t=F(P(I, \hat\theta_t))$ denote its corresponding objective value.  The reward for time step $t$ is
\begin{equation}
  \hat{R_t}=\min(O_{t}-F(P(I, \theta_t+p_t)),0).
\end{equation}
Thus, the model receives a reward proportional to the decrease in edge-straddling foreground pixels whenever the model updates the snap angle.
To speed up training, we use a variance-reduced version of the reward
$R_t = \hat{R_t}-b_t$ where $b_t$ is the average amount of decrease in pixel objectness coverage with a random policy at time $t$.

\vspace{-0.1in}
\section{Results}\label{sec:results}

Our results address \textbf{four main questions}: 
1) How efficiently can our approach identify the best snap angle? (Sec.~\ref{sec:pred_angle});
2) To what extent does the foreground ``pixel objectness'' objective properly capture objects important to human viewers? (Sec.~\ref{sec:preserve_object});
3) To what extent do human viewers favor snap-angle cubemaps over the default orientation? (Sec.~\ref{sec:human}); and 4) Might snap angles aid image recognition? (Sec.~\ref{sec:recognition}).

\vspace*{-0.1in}
\paragraph{Dataset}

We collect a dataset of 360$^{\circ}$ images to evaluate our approach; existing 360$^{\circ}$ datasets are topically narrow~\cite{xiao2012recognizing,su2016pano2vid,hu2017deep}, restricting their use for our goal.  
We \KG{use YouTube with the 360$^{\circ}$ filter to gather} videos from four activity categories---Disney, Ski, Parade, and Concert.  
After manually filtering out frames with only text or blackness, we have 150 videos and 14,076 total frames sampled at 1 FPS.

\vspace*{-0.1in}
\paragraph{Implementation details}

We implement our model with Torch, and optimize with stochastic gradient \KG{and REINFORCE}. We set the base learning rate to 0.01 and use momentum.  
\KG{We fix $A = 6.25\%$ for all results after visual inspection of a few human-taken cubemaps (not in the test set).}
See Supp. for all network architecture details.

\subsection{Efficient Snap Angle Prediction}\label{sec:pred_angle}

We first evaluate our snap angle prediction framework. 
We use all 14,076 frames, 75\% for training and 25\% for testing. We ensure testing and training data do \emph{not} come from the same video.
We define the following baselines:
\begin{itemize}
  \itemsep0em
  \item \textsc{Random rotate}:  Given a budget $T$, predict $T$ snap angles randomly (with no repetition).

    \item \textsc{Uniform rotate}: Given a budget $T$, predict $T$ snap angles uniformly sampled from all candidates.  
      When $T=1$, \textsc{Uniform} receives the \textsc{Canonical} view.  This is a strong baseline since it exploits the human recording bias in the starting view.  \KG{Despite the 360$^{\circ}$ range of the camera, photographers still tend to direct the ``front'' of the camera towards interesting content, in which case \textsc{Canonical} has some manual intelligence built-in.}
 
\item \bx{\textsc{Coarse-to-fine search}: 
Divide the search space into two uniform intervals and search the center snap angle in each interval. Then recursively search the better interval, until the budget is exhausted.}

\item \bx{\textsc{Pano2Vid(P2V)~\cite{su2016pano2vid}-adapted}:
We implement a snap angle variant inspired by the pipeline of Pano2Vid~\cite{su2016pano2vid}.
We replace C3D~\cite{tran2015learning} features (which require video) used in~\cite{su2016pano2vid} with F7 features from VGG~\cite{Simonyan14c} and train a logistic classifier to learn ``capture-worthiness"~\cite{su2016pano2vid} with Web images and randomly sampled panorama subviews (see Supp.). For a budget $T$, we evaluate $T$ ``glimpses" and choose the snap angle with the highest encountered capture-worthiness score.}  \KGthree{We stress that Pano2Vid addresses a different task: it creates a normal field-of-view video (discarding the rest) whereas we create a well-oriented omnidirectional image.  Nonetheless, we include this baseline to test their general approach of learning a framing prior from human-captured data.}

\item \textsc{Saliency}:  Select the angle that centers a cube face around the maximal saliency region.  Specifically, we compute the panorama's saliency map~\cite{liu2011learning} \BX{in equirectangular form} and blur it with a Gaussian kernel.  We then identify the $P\times P$ pixel square with the highest total saliency value, and predict the snap angle as the center of the square. Unlike the other methods, this baseline is not iterative, \KG{since the maximal saliency region does not change with rotations.}  We use a window size $P=30$.
\BX{Performance is not sensitive to $P$ for $20\leq P \leq 200$.}
\end{itemize}

\begin{figure}[t]
\centering
\renewcommand{\tabcolsep}{0pt}
\includegraphics[width=1\columnwidth]{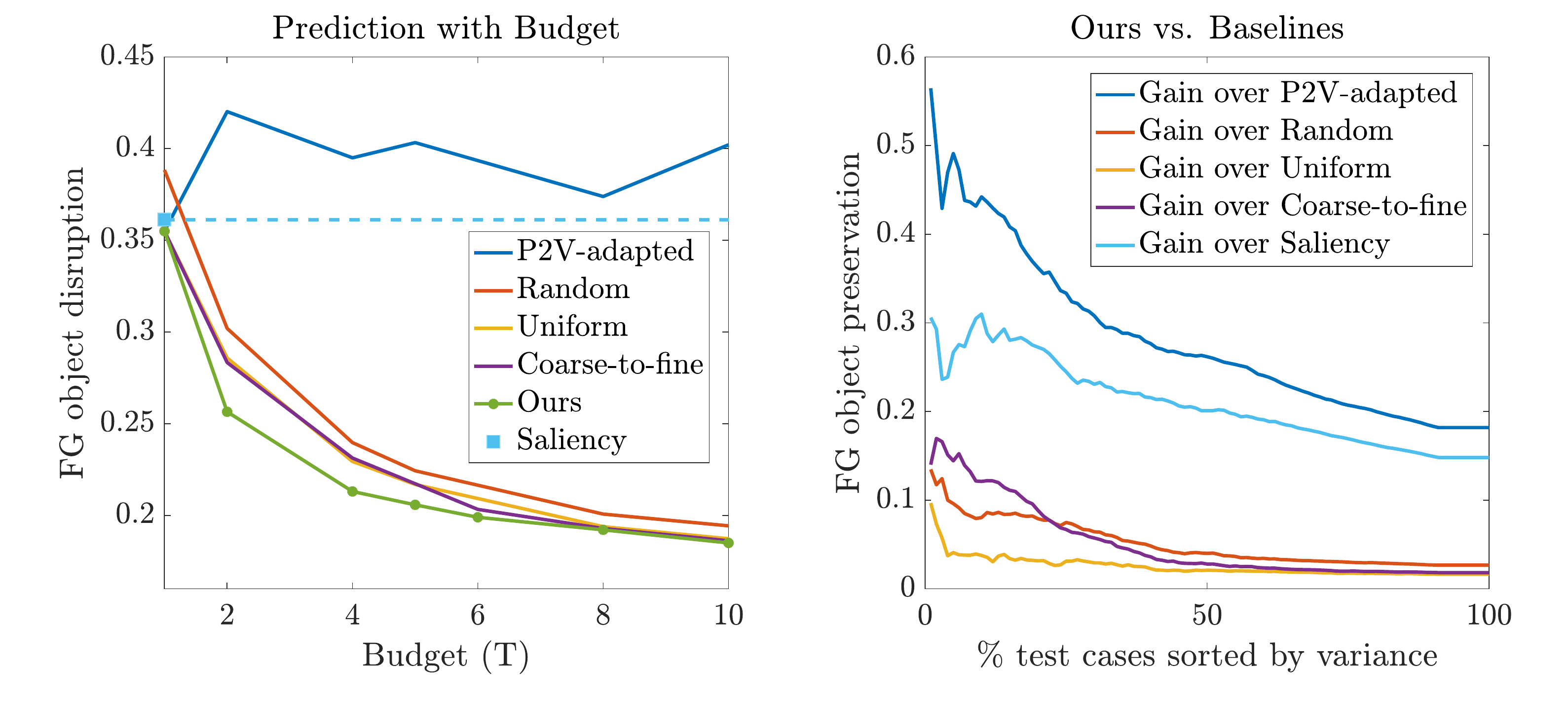}\vspace*{-0.2in}
\caption{Predicting snap angles in a timely manner.  Left: Given a budget, our method predicts snap angles with the least foreground disruption on cube edges.   \bo{Gains are larger for  smaller budgets, demonstrating our method's efficiency.}  Right: Our gain over the baselines \bx{(for a budget $T=4$)} as a function of the test cases' decreasing ``difficulty'', \cc{i.e., the variance in ground truth quality for candidate angles}.  See text.}
\label{fig:rl}
\vspace{-0.18in}
\end{figure}

We train our approach for a spectrum of budgets $T$, and report results in terms of \KG{the amount of foreground disruption} as a function of the budget.  Each unit of the budget corresponds to one round of rotating, re-rendering, and predicting foregrounds.  We score foreground disruption as the average  $F(P(I,\theta_t^*))$ across all four faces.

\bx{Figure~\ref{fig:rl} (left) shows the results. Our method achieves the least disruptions to foreground regions among all the competing methods.
\textsc{Uniform rotate} and \textsc{Coarse-to-fine search} perform better than \textsc{Random} because they benefit from hand-designed search heuristics.  
Unlike \textsc{Uniform rotate} and \textsc{Coarse-to-fine search}, our approach is content-based and learns to trigger different search strategies depending on what it observes.
When $T=1$, \textsc{Saliency} is better than \textsc{Random} but it underperforms our method and \textsc{Uniform}. 
\KGthree{\textsc{Saliency} likely has difficulty capturing important objects in panoramas, since the saliency model is trained with standard field-of-view images.  }
Directly adapting \textsc{Pano2Vid}~\cite{su2016pano2vid} for our problem results in unsatisfactory results.
A capture-worthiness classifier~\cite{su2016pano2vid} is relatively insensitive to the placement of important objects/people and therefore 
less suitable for the snap angle prediction task, which requires detailed modeling of object placement on \emph{all} faces of the cube.}

Figure~\ref{fig:rl} (right) plots our gains sorted by the test images' decreasing ``difficulty'' \bx{for a budget $T=4$}.  
In some test images, there is a \KG{high} variance, meaning certain snap angles are better than others.  However, for others,
\KGthree{all candidate rotations look similarly good}, in which case all methods will perform similarly. 
The righthand plot sorts the test images by their variance (in \KG{descending order}) in quality across all possible angles, and reports our method's gain as a function of that difficulty.  
Our method outperforms \textsc{P2V-adapted}, \textsc{Saliency}, \textsc{Coarse-to-fine search}, \textsc{Random} and \textsc{Uniform} by up to \bo{56\%, 31\%, 17\%, 14\% and 10\% (absolute),}
respectively.   Overall Figure~\ref{fig:rl} demonstrates that our method predicts the snap angle more efficiently than the baselines.
  
We have thus far reported efficiency in terms of abstract budget usage.
One unit of budget entails the following: projecting 
a typical panorama of size $960\times 1920$ pixels in equirectangular form to a cubemap (8.67 seconds with our Matlab implementation) and then computing pixel objectness (0.57 seconds). Our prediction method is very efficient and takes $0.003$ seconds to execute for a budget $T=4$ with a GeForce GTX 1080 GPU.  
Thus, for a budget $T=4$, the savings achieved by our method is approximately \bx{2.4 minutes (5x speedup)} per image compared to exhaustive search. \KGthree{Note that due to our method's efficiency, even if the Matlab projections were 1000x faster for all methods, our 5x speedup over the baseline would remain the same.}    Our method achieves a good tradeoff between speed and accuracy.

\vspace{-10pt}

\subsection{Justification for Foreground Object Objective}\label{sec:preserve_object}

Next we justify empirically the pixel objectness cube-edge objective.  To this end, we have human viewers identify important objects in the source panoramas, then evaluate to what extent our objective preserves them. 

Specifically, we randomly select 340 frames among those where: 1) Each frame is at least 10-seconds apart from the rest in order to ensure diversity in the dataset; 2) The difference in terms of overall pixel objectness between our method and the canonical view method is non-neglible.  
We collect annotations via Amazon Mechanical Turk.   Following the interface of~\cite{hu2017deep}, we present crowdworkers the panorama and instruct them to label any ``important objects'' with a bounding box---as many as they wish.
See Supp.~for interface and annotation statistics.

\bo{Here we consider \textsc{Pano2Vid(P2V)~\cite{su2016pano2vid}-adapted} and \textsc{Saliency} as defined in Sec.~\ref{sec:pred_angle} and two additional baselines:
1) \textsc{Canonical view}: produces a cubemap using the camera-provided orientation;
2) \textsc{Random view}: rotates the input panorama by an arbitrary angle and then generates the cubemap.}
\KGthree{Note that the other baselines in Sec.~\ref{sec:pred_angle} are not applicable here, since they are search mechanisms.}  

\BX{Consider the cube face $X$ that contains the largest number of foreground pixels from a given bounding box after projection.}
We evaluate the cubemaps of our method and the baselines based on the overlap score \KG{(IoU)} between
\BX{the foreground region from the cube face $X$ and the corresponding human-labeled important object,}
for each bounding box.  This metric is maximized when all pixels for the same object project to the same cube face; higher overlap indicates better preservation of important objects.

\begin{table}[t]
\centering
{\footnotesize \begin{tabular}{ |c|c|c|c|c|c||c|}
\hline

 & \textsc{Canonical} & \textsc{Random} & \textsc{Saliency}& \textsc{P2V-adapted}& \textsc{Ours} & \textsc{UpperBound}    \\ \hline

Concert  & 77.6\% & 73.9\% & 76.2\% & 71.6\% & \textbf{81.5\%} & 86.3\% \\ \hline
Ski      & 64.1\% & 72.5\% & 68.1\% & 70.1\% & \textbf{78.6\%} & 83.5\% \\ \hline
Parade   & 84.0\% & 81.2\% & 86.3\% & 85.7\% & \textbf{87.6\%} & 96.8\% \\ \hline
Disney   & 58.3\% & 57.7\% & 60.8\% & 60.8\% & \textbf{65.5\%} & 77.4\% \\ \hline
All      & 74.4\% & 74.2\% & 76.0\% & 75.0\% & \textbf{81.1\%} & 88.3\% \\ \hline
\end{tabular}}
\vspace{4pt}
\caption{Performance on preserving the integrity of objects explicitly identified as important by human observers.  Higher overlap scores are better. Our method outperforms \boc{all} baselines.} \vspace*{-0.2in}
\label{tab:imp_objects}

\vspace{-10pt}

\end{table}

Table~\ref{tab:imp_objects} shows the results.
Our method outperforms \bo{all} baselines by a large margin.
This supports our hypothesis that avoiding foreground objects along the cube edges helps preserve objects of interest to a viewer.  Snap angles achieve this goal much better than the baseline cubemaps.
The \textsc{UpperBound} corresponds to the maximum possible overlap achieved if exhaustively evaluating \emph{all} candidate angles, and helps gauge the difficulty of each category.   Parade and Disney have the highest and lowest upper bounds, respectively.
In Disney images, the camera is often carried by the recorders, so important objects/persons appear relatively large in the panorama and cannot fit in a single cube face, hence a lower upper bound score.
On the contrary, in Parade images the camera is often placed in the crowd and far away from important objects, so each can be confined to a single face.
The latter also explains why the baselines do best (though still weaker than ours) on Parade images. 
\boc{An ablation study decoupling the pixel objectness performance from snap angle performance pinpoints the effects of foreground quality on our approach (see Supp.).}

\vspace{-10pt}
\subsection{User Study: Perceived Quality}\label{sec:human}

\begin{table}[t]
\centering

{\footnotesize \begin{tabular}{ cccc||ccc}
\toprule

& Prefer \textsc{Ours} &Tie& Prefer \textsc{Canonical} & Prefer \textsc{Ours} &Tie& Prefer \textsc{Random}  \\ \midrule
Parade & 54.8\%& 16.5\% &  28.7\%   & 70.4\%& 9.6\% &  20.0\%\\
Concert &48.7\%&16.2\%  &35.1\%     &52.7\%&16.2\%  &31.1\% \\
Disney &44.8\%&17.9\%  &37.3\%      &72.9\%&8.5\%  &18.6\%\\
Ski &64.3\%&8.3\%  &27.4\%          &62.9\%&16.1\%  &21.0\%\\
All &53.8\%&14.7\%&31.5\%           &65.3\%&12.3\%&22.4\%\\
\bottomrule

\end{tabular}}
\vspace{4pt}
\caption{User study result comparing cubemaps outputs for perceived quality. \bx{Left: Comparison between our method and \textsc{Canonical}. Right: Comparison between our method and \textsc{Random}.}}

\label{tab:human_study}
\vspace{-15pt}
\end{table}

\begin{table}[t]

\centering
{\scriptsize \begin{tabular}{ |c|c|c|c|c|c|}
\hline

   & Concert           & Ski              & Parade                  & Disney                         &  All (normalized)    \\ \hline
\multicolumn{6}{|c|}{Image Memorability~\cite{ICCV15_Khosla}} \\ \hline 
\textsc{Canonical} & \textbf{71.58}    & 69.49          & 67.08          & 70.53                 & 46.8\%                 \\ \hline
\textsc{Random}    & 71.30             & 69.54          & 67.27          &70.65                 &  48.1\%                 \\ \hline 
\textsc{Saliency} & 71.40    & 69.60          & 67.35         & 70.58                 & 49.9\%                 \\ \hline
\textsc{P2V-adapted}    & 71.34             & 69.85          & 67.44          &70.54                 &  52.1\%                 \\ \hline
\textsc{Ours}      & 71.45             & \textbf{70.03} & \textbf{67.68} & \textbf{70.87}        &  \textbf{59.8\%} \\ \hline\hline
\textsc{Upper}     & 72.70             & 71.19          & 68.68          & 72.15                 &  --        \\ \hline
\multicolumn{6}{|c|}{Image  Aesthetics~\cite{kong2016aesthetics}}  \\ \hline
\textsc{Canonical} & 33.74             & 41.95           & 30.24          & 32.85                 &  44.3\% \\ \hline
\textsc{Random}    & 32.46             & 41.90           & 30.65          & 32.79                 &  42.4\% \\ \hline
\textsc{Saliency} & 34.52    & 41.87          & 30.81          & 32.54                 & 47.9\%                 \\ \hline
\textsc{P2V-adapted}    & 34.48             & 41.97          & 30.86          &\textbf{33.09}                 &  48.8\%                 \\ \hline
\textsc{Ours}      & \textbf{35.05}    & \textbf{42.08}  & \textbf{31.19}  & 32.97      &  \textbf{52.9\%} \\ \hline\hline
\textsc{Upper}     & 38.45                  & 45.76                & 34.74               & 36.81                      &  --    \\ \hline
\end{tabular}}

\vspace{4pt}
\caption{Memorability and aesthetics scores.}

\label{tab:img_pro}
\vspace{-25pt}
\end{table}

\begin{figure*}[t!]
\centering
\renewcommand{\tabcolsep}{0pt}
\includegraphics[width=1\columnwidth]{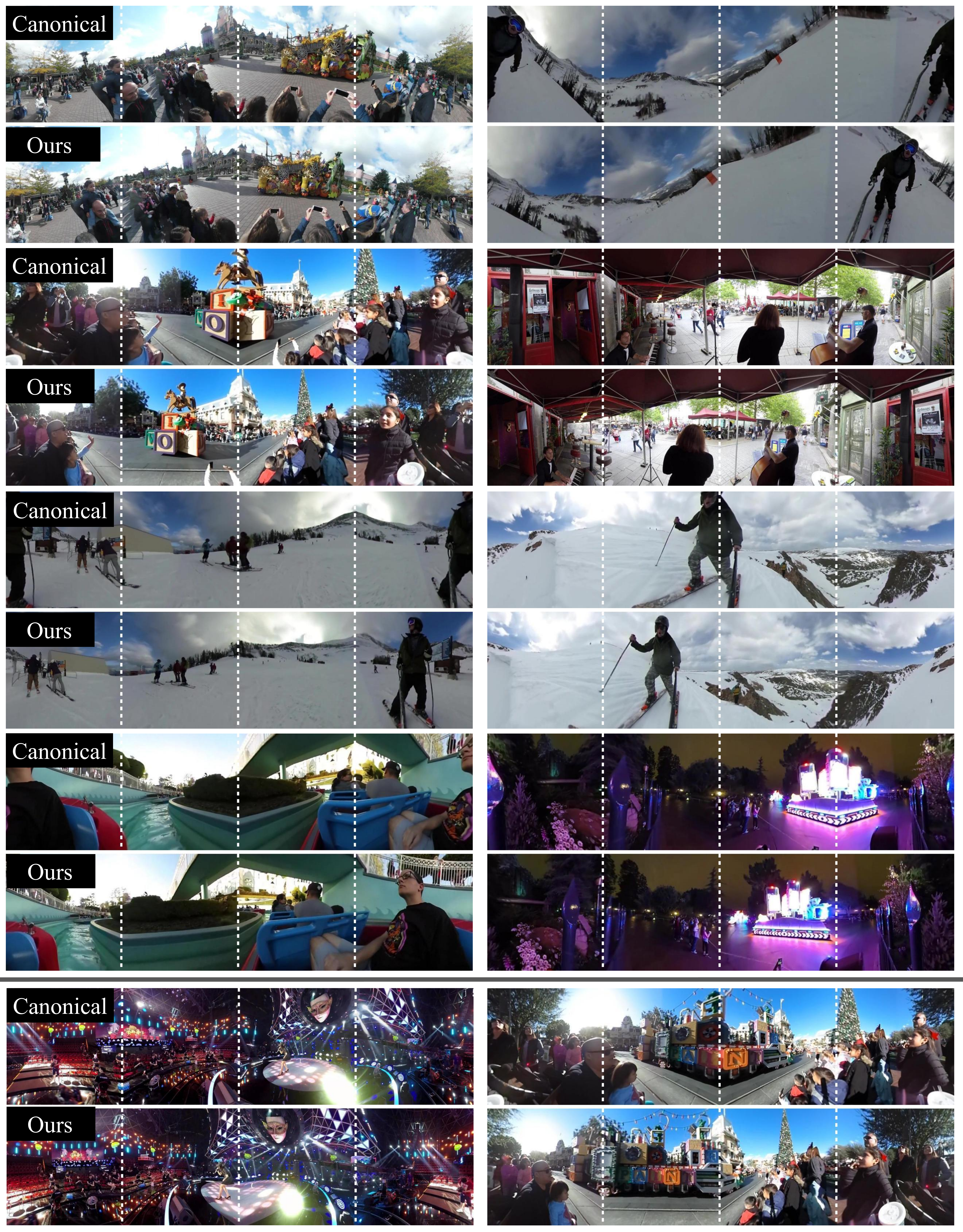}
\caption{Qualitative examples of default \textsc{Canonical} cubemaps and our snap angle cubemaps.  Our method produces cubemaps that place important objects/persons in the same cube face to preserve the foreground integrity. Bottom two rows show failure cases. In the bottom left, pixel objectness~\cite{jain2017pixel} does not recognize the round stage as foreground, and therefore our method splits the stage onto two different cube faces, creating a distorted heart-shaped stage. \boc{In the bottom right, the train is too large to fit in a single cube.}}
\label{fig:qual}
\end{figure*}

\KGthree{Having justified the perceptual relevance of the cube-edge foreground objective (Sec.~\ref{sec:preserve_object})}, next we perform a user study to gauge perceptual quality of our results.  Do snap angles produce cube faces that look like human-taken photos?
We evaluate on the same image set used in Sec.~\ref{sec:preserve_object}.

We present cube faces produced by our method and one of the baselines at a time in arbitrary order and inform subjects the two sets are photos from the same scene but taken by different photographers. We instruct them to consider composition and viewpoint in order to decide which set of photos is more pleasing (see Supp.).  
To account for the subjectivity of the task, we issue each sample to 5 distinct workers and aggregate responses with majority vote.  $98$ unique MTurk crowdworkers participated in the study.

\bx{Table~\ref{tab:human_study} shows the results. Our method outperforms the \textsc{Canonical} baseline by more than 22\% and the \textsc{Random} baseline by 42.9\%.} 
This result supports our claim that by preserving object integrity, our method produces cubemaps that align better with human perception of quality photo composition.  
Figure~\ref{fig:qual} shows qualitative examples.  As shown in the first two examples (top two rows), our method is able to place an important person in the same cube face whereas the baseline splits each person and projects a person onto two cube faces.  We also present two failure cases in the last two rows. In the bottom left, pixel objectness does not recognize the stage as foreground, and therefore our method places the stage on two different cube faces, creating a distorted heart-shaped stage. \boc{Please see Supp. for pixel objectness map input for failure cases.}

So far, Table~\ref{tab:imp_objects} confirms empirically that our foreground-based objective does preserve those objects human viewers deem important, and Table~\ref{tab:human_study} shows that human viewers have an absolute preference for snap angle cubemaps over other projections. 
As a final test of snap angle cubemaps' perceptual quality, we score them using state-of-the-art metrics for \emph{aesthetics}~\cite{kong2016aesthetics} and \emph{memorability}~\cite{ICCV15_Khosla}.
Since both models are trained on images annotated by people (for their aesthetics and memorability, respectively), higher scores indicate higher correlation with these perceived properties (though of course no one learned metric can perfectly represent human opinion).

Table~\ref{tab:img_pro} shows the results.
We report the raw scores $s$ per class as well as the score over all classes, normalized as $\frac{s-s_{min}}{s_{max}-s_{min}}$, where $s_{min}$ and $s_{max}$ denote the lower and upper bound, respectively.
Because the metrics are fairly tolerant to local rotations, there is a limit to how well they can capture subtle differences in cubemaps.  Nonetheless, our method outperforms the baselines overall. 
\boc{Given these metrics' limitations, the user study in Table~\ref{tab:human_study} offers the most direct and conclusive evidence for snap angles' perceptual advantage. }

\vspace{-5pt}

\subsection{Cubemap Recognition from Pretrained Nets}\label{sec:recognition}

\begin{table}[t]\centering
\begin{tabular}{c|ccc}\toprule
 &   \textsc{Canonical}   &   \textsc{Random}       &   \textsc{Ours} \\ \midrule
Single  &   68.5       & 69.4         &  \textbf{70.1}        \\
Pano    &   66.5       & 67.0         & \textbf{ 68.1}         \\
\bottomrule
\end{tabular}
\vspace{4pt}
\caption{Image recognition accuracy (\%). Snap angles help align the 360$^{\circ}$ data's statistics with that of normal FOV Web photos, enabling easier transfer from conventional pretrained networks.}
\label{tab:reg_google}
\vspace{-25pt}
\end{table}

Since snap angles provide projections that better mimic human-taken photo composition, we hypothesize that they also align better with conventional FOV images, compared to cubemaps in their canonical orientation.  This suggests that snap angles may better align with Web photos (typically used to train today's recognition systems), which in turn could help standard recognition models perform well on 360$^{\circ}$ panoramas.  We present a preliminary proof-of-concept experiment to test this hypothesis.

We train a multi-class CNN classifier to distinguish the four activity categories in our 360$^{\circ}$ dataset (Disney, Parade, etc.).  The classifier uses ResNet-101~\cite{he2016deep} pretrained on ImageNet~\cite{russakovsky2015imagenet} and fine-tuned on 300 training images per class downloaded from Google Image Search (see Supp.).
Note that in all experiments until now, the category labels on the 360$^{\circ}$ dataset were invisible to our algorithm.
We randomly select 250 panoramas per activity as a test set.   Each panorama is projected to a cubemap with the different projection methods, and we compare the resulting recognition rates.

Table~\ref{tab:reg_google} shows the results.  We report recognition accuracy in two forms: \emph{Single}, which 
treats each individual cube face as a test instance, and \emph{Pano}, which classifies the entire panorama by multiplying the predicted posteriors from all cube faces.  For both cases, snap angles produce cubemaps that achieve the best recognition rate. 
\boc{This result hints at the potential for snap angles to be a bridge between pretrained normal FOV networks on the one hand and 360$^{\circ}$ images on the other hand. \KGtwo{That said, the margin is slim, and} the full impact of snap angles for recognition warrants further exploration.  }

\vspace{-5pt}
\section{Conclusions}
\vspace{-5pt}

We introduced the snap angle prediction problem for rendering 360$^{\circ}$ images. In contrast to previous work that assumes either a fixed or manually supplied projection angle, we propose to automatically predict the angle that will best preserve detected foreground objects. \boc{We present a framework 
to efficiently and accurately predict the  snap angle in novel panoramas.} 
\boc{We demonstrate the advantages of the proposed method, both in terms of human perception and several statistical metrics.  Future work will explore ways to generalize snap angles to video data and expand snap angle prediction to other projection models.}

\vspace{-5pt}
\paragraph{Acknowledgements} \boc{This research is supported in part by NSF IIS-1514118 and a Google Faculty Research Award. We also gratefully acknowledge a GPU donation from Facebook.}

\clearpage

\bibliographystyle{splncs}
\bibliography{snapAngle}

\begin{thebibliography}{10}

\bibitem{su2016pano2vid}
Su, Y.C., Jayaraman, D., Grauman, K.:
\newblock Pano2vid: Automatic cinematography for watching 360° videos.
\newblock In: ACCV. (2016)

\bibitem{su2017making}
Su, Y.C., Grauman, K.:
\newblock Making 360° video watchable in 2d: Learning videography for click
  free viewing.
\newblock In: CVPR. (2017)

\bibitem{hu2017deep}
Hu, H.N., Lin, Y.C., Liu, M.Y., Cheng, H.T., Chang, Y.J., Sun, M.:
\newblock Deep 360 pilot: Learning a deep agent for piloting through 360◦
  sports videos.
\newblock In: CVPR. (2017)

\bibitem{lai2017semantic}
Lai, W.S., Huang, Y., Joshi, N., Buehler, C., Yang, M.H., Kang, S.B.:
\newblock Semantic-driven generation of hyperlapse from 360° video.
\newblock IEEE Transactions on Visualization and Computer Graphics (2017)

\bibitem{snyder1997flattening}
Snyder, J.P.:
\newblock Flattening the earth: two thousand years of map projections.
\newblock University of Chicago Press (1997)

\bibitem{greene1986environment}
Greene, N.:
\newblock Environment mapping and other applications of world projections.
\newblock IEEE Computer Graphics and Applications (1986)

\bibitem{sharpless2010pannini}
Sharpless, T.K., Postle, B., German, D.M.:
\newblock Pannini: a new projection for rendering wide angle perspective
  images.
\newblock In: International Conference on Computational Aesthetics in Graphics,
  Visualization and Imaging. (2010)

\bibitem{kim2017automatic}
Kim, Y.W., Jo, D.Y., Lee, C.R., Choi, H.J., Kwon, Y.H., Yoon, K.J.:
\newblock Automatic content-aware projection for 360° videos.
\newblock In: ICCV. (2017)

\bibitem{li2015geodesic}
Li, D., He, K., Sun, J., Zhou, K.:
\newblock A geodesic-preserving method for image warping.
\newblock In: CVPR. (2015)

\bibitem{carroll2009optimizing}
Carroll, R., Agrawala, M., Agarwala, A.:
\newblock Optimizing content-preserving projections for wide-angle images.
\newblock In: ACM Transactions on Graphics. (2009)

\bibitem{tehrani2016correcting}
Tehrani, M.A., Majumder, A., Gopi, M.:
\newblock Correcting perceived perspective distortions using object specific
  planar transformations.
\newblock In: ICCP. (2016)

\bibitem{carroll2010image}
Carroll, R., Agarwala, A., Agrawala, M.:
\newblock Image warps for artistic perspective manipulation.
\newblock In: ACM Transactions on Graphics. (2010)

\bibitem{kopf2009locally}
Kopf, J., Lischinski, D., Deussen, O., Cohen-Or, D., Cohen, M.:
\newblock Locally adapted projections to reduce panorama distortions.
\newblock In: Computer Graphics Forum, Wiley Online Library (2009)

\bibitem{wang2015panorama}
Wang, Z., Jin, X., Xue, F., He, X., Li, R., Zha, H.:
\newblock Panorama to cube: a content-aware representation method.
\newblock In: SIGGRAPH Asia Technical Briefs. (2015)

\bibitem{fb2015cubemap}

\newblock
  \url{https://code.facebook.com/posts/1638767863078802/under-the-hood-building-360-video/}

\bibitem{google2017eac}

\newblock
  \url{https://www.blog.google/products/google-vr/bringing-pixels-front-and-center-vr-video/}

\bibitem{kong2016aesthetics}
Kong, S., Shen, X., Lin, Z., Mech, R., Fowlkes, C.:
\newblock Photo aesthetics ranking network with attributes and content
  adaptation.
\newblock In: ECCV. (2016)

\bibitem{isola2011makes}
Isola, P., Xiao, J., Torralba, A., Oliva, A.:
\newblock What makes an image memorable?
\newblock In: CVPR. (2011)

\bibitem{xiong2014detecting}
Xiong, B., Grauman, K.:
\newblock Detecting snap points in egocentric video with a web photo prior.
\newblock In: ECCV. (2014)

\bibitem{gygli2013interestingness}
Gygli, M., Grabner, H., Riemenschneider, H., Nater, F., Van~Gool, L.:
\newblock The interestingness of images.
\newblock In: ICCV. (2013)

\bibitem{ICCV15_Khosla}
Khosla, A., Raju, A.S., Torralba, A., Oliva, A.:
\newblock Understanding and predicting image memorability at a large scale.
\newblock In: ICCV. (2015)

\bibitem{chang2013rectangling}
Chang, C.H., Hu, M.C., Cheng, W.H., Chuang, Y.Y.:
\newblock Rectangling stereographic projection for wide-angle image
  visualization.
\newblock In: ICCV. (2013)

\bibitem{zelnik2005squaring}
Zelnik-Manor, L., Peters, G., Perona, P.:
\newblock Squaring the circle in panoramas.
\newblock In: ICCV. (2005)

\bibitem{kopf2007capturing}
Kopf, J., Uyttendaele, M., Deussen, O., Cohen, M.F.:
\newblock Capturing and viewing gigapixel images.
\newblock In: ACM Transactions on Graphics. (2007)

\bibitem{mnih2014recurrent}
Mnih, V., Heess, N., Graves, A., Kavukcuoglu, K.:
\newblock Recurrent models of visual attention.
\newblock In: NIPS. (2014)

\bibitem{caicedo2015active}
Caicedo, J.C., Lazebnik, S.:
\newblock Active object localization with deep reinforcement learning.
\newblock In: ICCV. (2015)

\bibitem{mathe2016reinforcement}
Mathe, S., Pirinen, A., Sminchisescu, C.:
\newblock Reinforcement learning for visual object detection.
\newblock In: CVPR. (2016)

\bibitem{jayaraman2016look}
Jayaraman, D., Grauman, K.:
\newblock Look-ahead before you leap: End-to-end active recognition by
  forecasting the effect of motion.
\newblock In: ECCV. (2016)

\bibitem{jayaraman2017learning}
Jayaraman, D., Grauman, K.:
\newblock Learning to look around: Intelligently exploring unseen environments
  for unknown tasks.
\newblock In: CVPR. (2018)

\bibitem{jayaraman2018end}
Jayaraman, D., Grauman, K.:
\newblock End-to-end policy learning for active visual categorization.
\newblock PAMI (2018)

\bibitem{yeung2016end}
Yeung, S., Russakovsky, O., Mori, G., Fei-Fei, L.:
\newblock End-to-end learning of action detection from frame glimpses in
  videos.
\newblock In: CVPR. (2016)

\bibitem{alwassel2017action}
Alwassel, H., Heilbron, F.C., Ghanem, B.:
\newblock Action search: Learning to search for human activities in untrimmed
  videos.
\newblock arXiv preprint arXiv:1706.04269 (2017)

\bibitem{singh2016multi}
Singh, B., Marks, T.K., Jones, M., Tuzel, O., Shao, M.:
\newblock A multi-stream bi-directional recurrent neural network for
  fine-grained action detection.
\newblock In: CVPR. (2016)

\bibitem{su2016leaving}
Su, Y.C., Grauman, K.:
\newblock Leaving some stones unturned: dynamic feature prioritization for
  activity detection in streaming video.
\newblock In: ECCV. (2016)

\bibitem{jain2017pixel}
Xiong, B., Jain, S.D., Grauman, K.:
\newblock Pixel objectness: Learning to segment generic objects automatically
  in images and videos.
\newblock PAMI (2018)

\bibitem{zitnick2014edge}
Zitnick, C.L., Doll{\'a}r, P.:
\newblock Edge boxes: Locating object proposals from edges.
\newblock In: ECCV. (2014)

\bibitem{carreira2012cpmc}
Carreira, J., Sminchisescu, C.:
\newblock C{P}{M}{C}: Automatic object segmentation using constrained
  parametric min-cuts.
\newblock PAMI (2011)

\bibitem{jiang2013}
Jiang, P., Ling, H., Yu, J., Peng, J.:
\newblock Salient region detection by ufo: Uniqueness, focusness, and
  objectness.
\newblock In: ICCV. (2013)

\bibitem{pinheiro2015learning}
Pinheiro, P.O., Collobert, R., Doll{\'a}r, P.:
\newblock Learning to segment object candidates.
\newblock In: NIPS. (2015)

\bibitem{liu2011learning}
Liu, T., Yuan, Z., Sun, J., Wang, J., Zheng, N., Tang, X., Shum, H.Y.:
\newblock Learning to detect a salient object.
\newblock PAMI (2011)

\bibitem{dhar2011high}
Dhar, S., Ordonez, V., Berg, T.L.:
\newblock High level describable attributes for predicting aesthetics and
  interestingness.
\newblock In: CVPR. (2011)

\bibitem{williams1992simple}
Williams, R.J.:
\newblock Simple statistical gradient-following algorithms for connectionist
  reinforcement learning.
\newblock Machine learning (1992)

\bibitem{xiao2012recognizing}
Xiao, J., Ehinger, K.A., Oliva, A., Torralba, A.:
\newblock Recognizing scene viewpoint using panoramic place representation.
\newblock In: CVPR. (2012)

\bibitem{tran2015learning}
Tran, D., Bourdev, L., Fergus, R., Torresani, L., Paluri, M.:
\newblock Learning spatiotemporal features with 3d convolutional networks.
\newblock In: ICCV. (2015)

\bibitem{Simonyan14c}
Simonyan, K., Zisserman, A.:
\newblock Very deep convolutional networks for large-scale image recognition.
\newblock CoRR \textbf{abs/1409.1556} (2014)

\bibitem{he2016deep}
He, K., Zhang, X., Ren, S., Sun, J.:
\newblock Deep residual learning for image recognition.
\newblock In: CVPR. (2016)

\bibitem{russakovsky2015imagenet}
Russakovsky, O., Deng, J., Su, H., Krause, J., Satheesh, S., Ma, S., Huang, Z.,
  Karpathy, A., Khosla, A., Bernstein, M.,  et~al.:
\newblock Imagenet large scale visual recognition challenge.
\newblock IJCV (2015)

\bibitem{profx}

\newblock \url{https://github.com/jianxiongxiao/ProfXkit}

\bibitem{5432215}
Liu, T., Yuan, Z., Sun, J., Wang, J., Zheng, N., Tang, X., Shum, H.Y.:
\newblock Learning to detect a salient object.
\newblock IEEE Transactions on Pattern Analysis and Machine Intelligence (2011)

\bibitem{jiang2013saliency}
Jiang, B., Zhang, L., Lu, H., Yang, C., Yang, M.H.:
\newblock Saliency detection via absorbing markov chain.
\newblock In: ICCV. (2013)

\end{thebibliography}

\section{Supplementary Material}

\subsection{Justification for predicting snap angle in azimuth only}

This section accompanies Sec. 3.1 in the main paper.

As discussed in the paper, views from a horizontal camera position (elevation 0$^{\circ}$) are typically more plausible than other elevations due to the human recording bias.  The human photographer typically holds the camera with the ``top" to the sky/ceiling.  

\begin{figure*}[t!]
\centering
\renewcommand{\tabcolsep}{0pt}
\includegraphics[width=1\columnwidth]{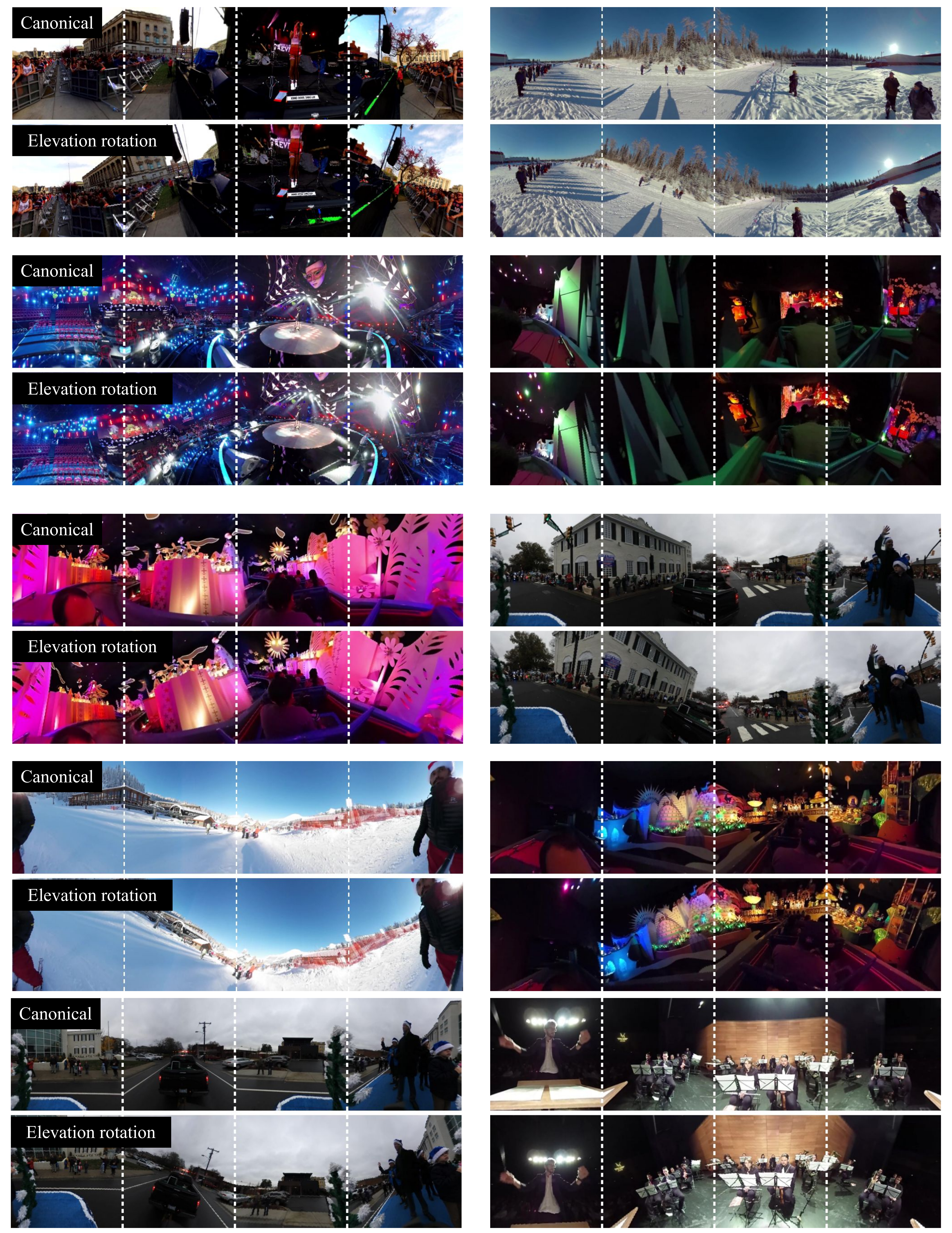}

\caption{Example of cubemaps when rotating in elevation. Views from a horizontal camera position (elevation 0$^{\circ}$) are more informative than others due to the natural human recording bias. In addition, rotation in elevation often makes cubemap faces appear tilted (e.g., building in the second row).  Therefore, we optimize for the azimuth snap angle only.}
\label{fig:supp_elevation}
\end{figure*}

Figure~\ref{fig:supp_elevation} shows examples of rotations in elevation for several panoramas.
We see that the recording bias makes the 0 (canonical) elevation fairly good as it is.  In contrast, rotation in elevation often makes the cubemaps appear tilted (e.g., the building in the second row). Without loss of generality, we focus on snap angles in azimuth only, and jointly
optimize the front/left/right/back faces of the cube.

\subsection{Details on network architecture}

This section accompanies Sec. 3.2 in the main paper.

\begin{figure*}[t]
\centering
\renewcommand{\tabcolsep}{0pt}
\includegraphics[width=0.95\columnwidth]{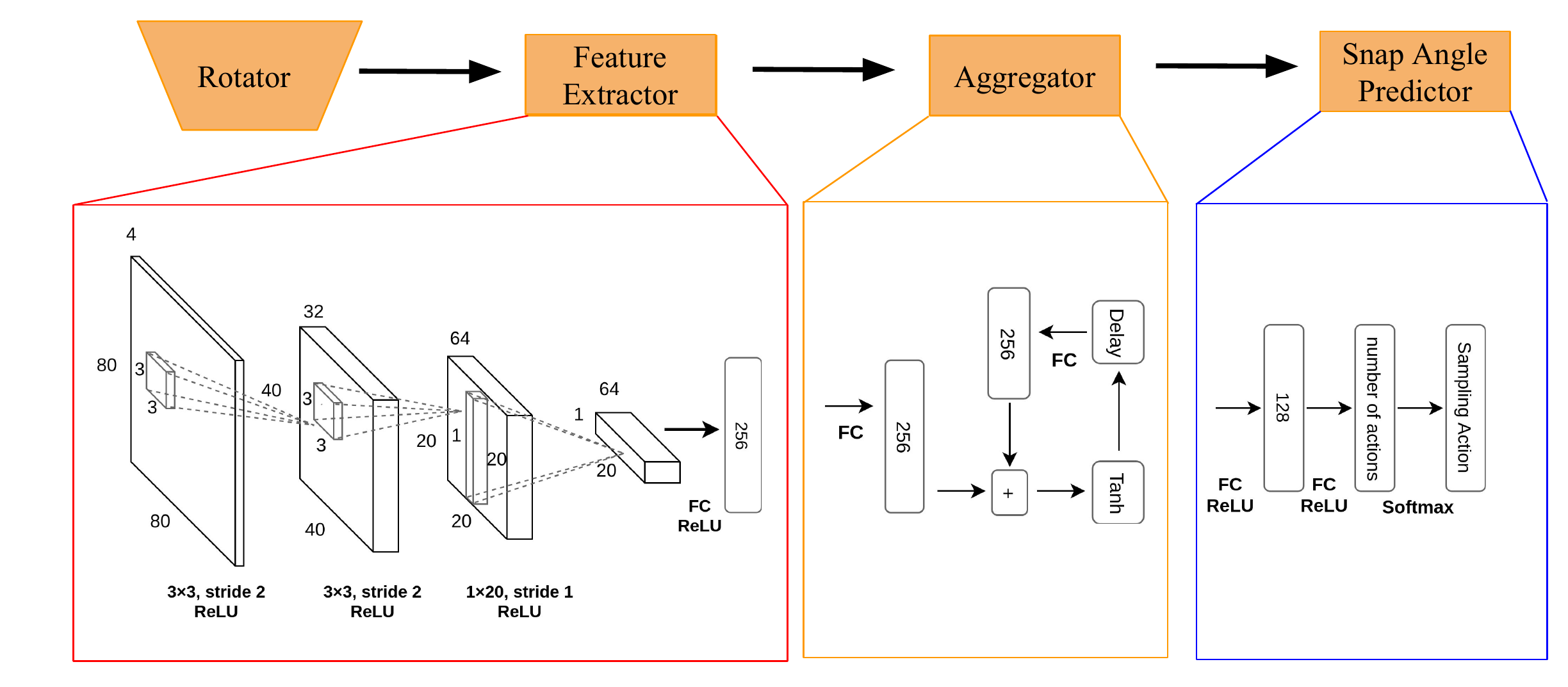}
\caption{A detailed diagram showing our network architecture. In the top, the small schematic shows the connection between each network module. Then we present the details of each module in the bottom. Our network proceeds from left to right. The \textit{feature extractor} consists of a sequence of convolutions (with kernel size and convolution stride written under the diagram) followed by a fully connected layer. In the bottom, ``FC'' denotes a fully connected layer and ``ReLU'' denotes a rectified linear unit. The \textit{aggregator} is a recurrent neural network. The ``Delay'' layer stores its current internal hidden state and outputs them at the next time step. In the end, the \textit{predictor} samples an action stochastically based on the multinomial pdf from the Softmax layer.}
\label{fig:supp_net}
\end{figure*}

Recall that our framework consists of four modules: a \textit{rotator}, a \textit{feature extractor}, an \textit{aggregator}, and a \textit{snap angle predictor}. At each time step, it processes the data and produces a cubemap (\textit{rotator}), extracts learned features (\textit{feature extractor}), integrates information over time (\textit{aggregator}), and predicts the next snap angle (\textit{snap angle predictor}). We show the details of each module in Figure~\ref{fig:supp_net}.

\subsection{Training details for \textsc{Pano2Vid(P2V)~\cite{su2016pano2vid}-adapted}}\label{sec:pano}

This section accompanies Sec 4.1 in the main paper.

For each of the activity categories in our 360$^{\circ}$ dataset (Disney, Parade, etc.), 
we query Google Image Search engine and then manually filter out irrelevant images. We obtain about $300$ images for each category.   
We use the Web images as positive training samples and randomly sampled panorama subviews as negative training sampling.

\clearpage

\subsection{Interface for collecting important objects/persons}
 
This section accompanies Sec. 4.2 in the main paper.

We present the interface for Amazon Mechanical Turk that is used to collect important objects and persons. The interface is developed based on~\cite{profx}. We present crowdworkers the panorama and instruct them to label any important objects with a bounding box---as many as they wish. A total of $368$ important objects/persons are labeled. The maximum number of labeled important objects/persons for a single image is $7$.

\begin{figure*}[t]
\centering
\renewcommand{\tabcolsep}{0pt}
\includegraphics[width=0.95\columnwidth]{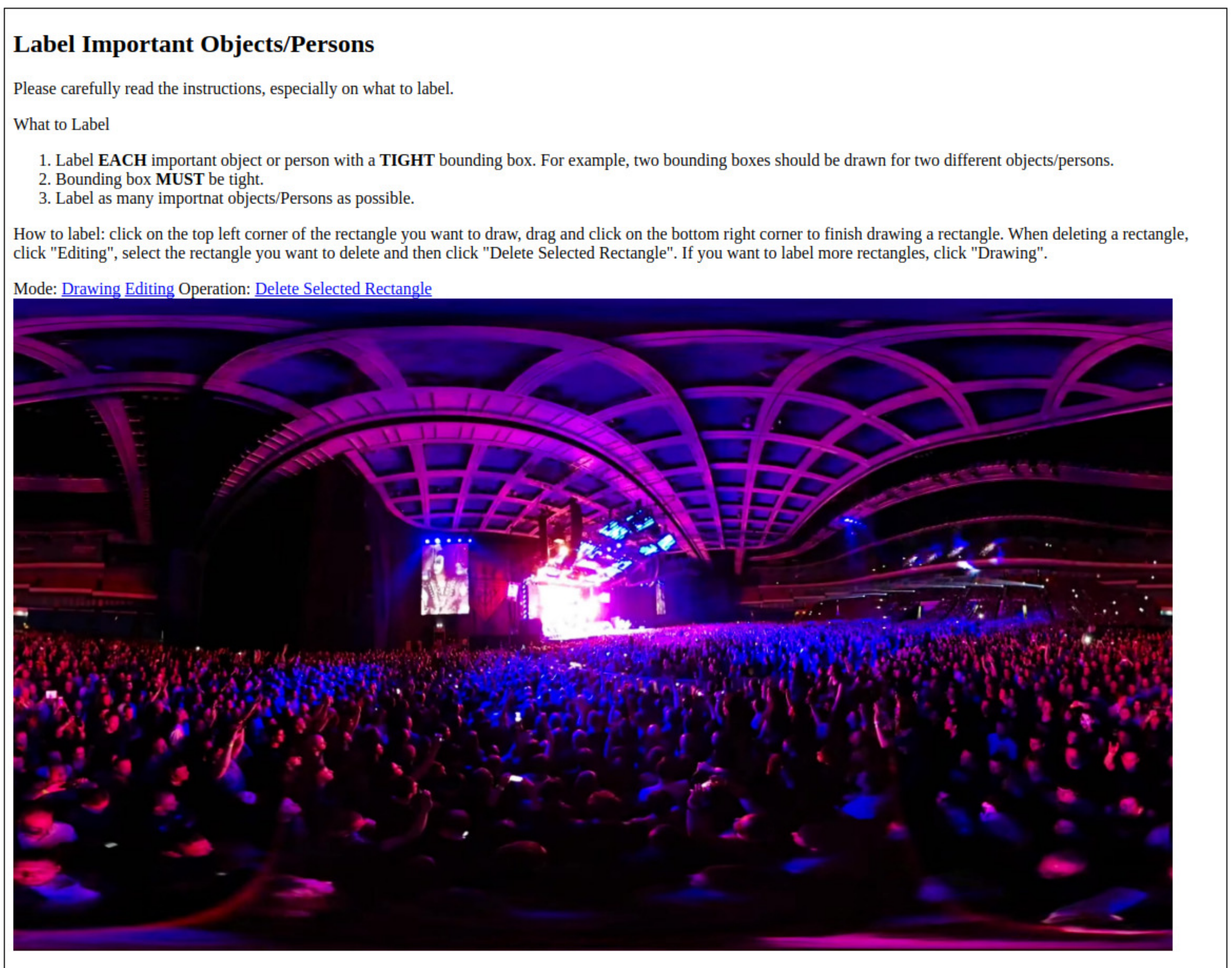}
\caption{Interface for Amazon Mechanical Turk when collecting important objects/persons.
We present crowdworkers the panorama and instruct them to label any important objects with a bounding box---as many as they wish.}
\label{fig:supp_important}
\end{figure*}

\subsection{More results when ground-truth objectness maps are used as input}
\bo{This section accompanies Sec. 4.2 in the main paper.}

\bo{We first got manual labels for the pixel-wise foreground for 50 randomly selected panoramas ($200$ faces). The IoU score between the pixel objectness prediction and ground truth is 0.66 with a recall of 0.82, whereas alternate foreground methods we tried \cite{5432215,jiang2013saliency} obtain only 0.35-0.40 IoU and 0.67-0.74 recall  on the same data.}
 
 \bo{We then compare results when either ground-truth foreground or pixel objectness is used as input (without re-training our model) and report the foreground disruption with respect to the ground-truth. ``Best possible" is the oracle result. Pixel objectness serves as a good proxy for ground-truth foreground maps. Error decreases with each rotation. Table~\ref{tab:gt} pinpoints to what extent having even better foreground inputs would also improve snap angles.}

 \begin{table}[t]
\centering
 \begin{tabular}{ |c|c|c|c|c|c|c|c|}
\hline

Budget (T)  & 2 & 4 & 6 & 8  & 10 & Best Possible & \textsc{Canonical}    \\ \hline
Ground truth input & 0.274 & 0.259 & 0.248 & 0.235 & 0.234 & 0.231 & 0.352\\ \hline
Pixel objectness input & 0.305 & 0.283 & 0.281 & 0.280  & 0.277 & 0.231 & 0.352\\ \hline

\end{tabular}

\vspace{4pt}
\caption{Decoupling pixel objectness and snap angle performance: 
 error (lower is better) when ground truth or pixel objectness is used as input.}
\label{tab:gt}
\end{table}

\subsection{Interface for user study on perceived quality}

This section accompanies Sec. 4.3 in the main paper.

\begin{figure*}[t]
\centering
\renewcommand{\tabcolsep}{0pt}
\includegraphics[width=0.95\columnwidth]{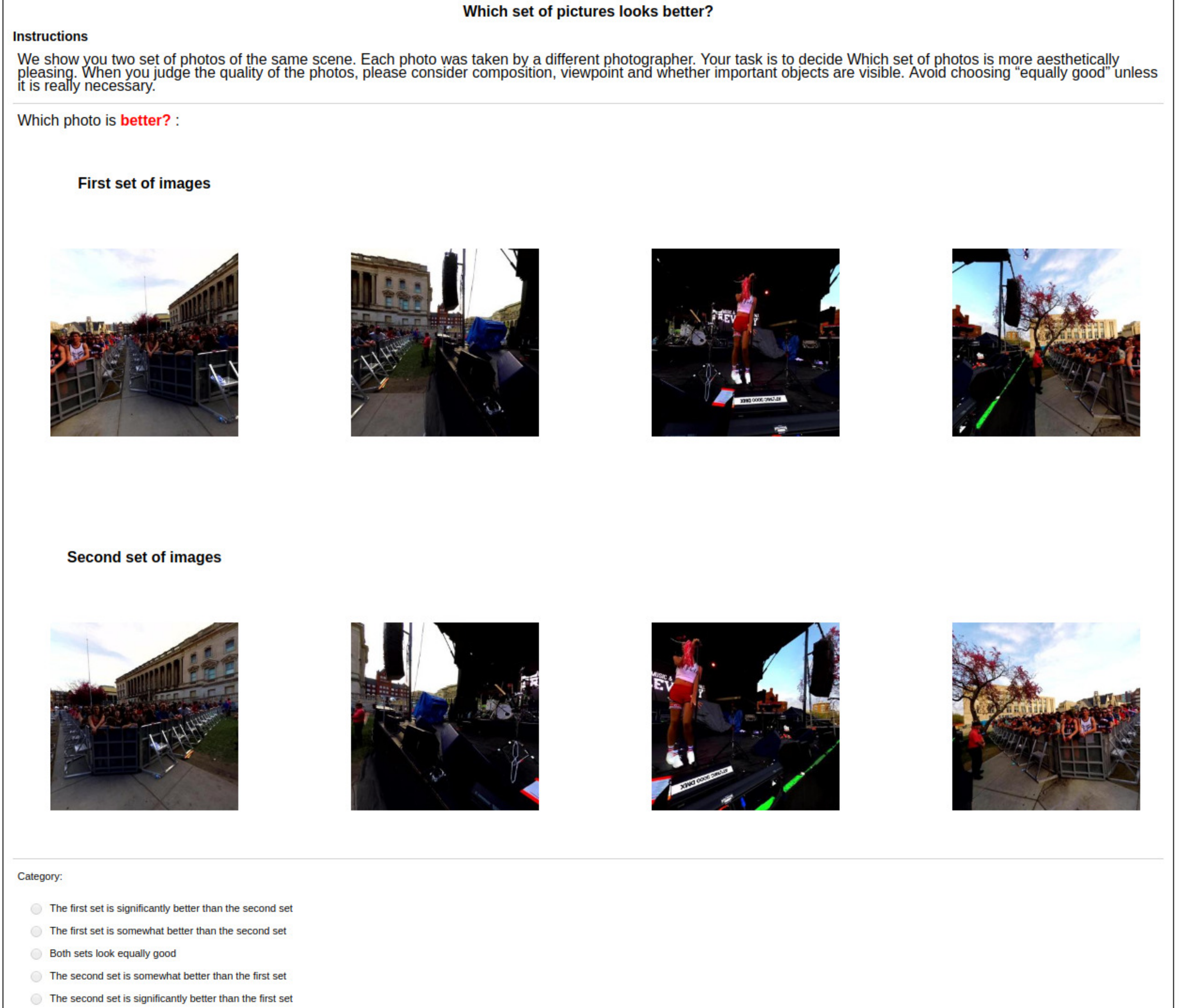}
\caption{Interface for user study on perceived quality. Workers were required to rate each image into one of five categories. We issue each sample to 5 distinct workers.}
\label{fig:supp_human}
\end{figure*}

We present the interface for the user study on perceived quality in Figure~\ref{fig:supp_human}. Workers were required to rate each image into one of five categories: (a) The first set is significantly better, (b) The first set is somewhat better, (c) Both sets look similar, (d) The second set is somewhat better and (e) The second set is significantly better. We also instruct them to avoid to choose option (c) unless it is really necessary. Since the task can be ambiguous and subjective, we issued each task to 5 distinct workers. Every time a comparison between the two sets receives a rating of category (a), (b), (c), (d) or (e) from any of the 5 workers, it receives 2, 1, 0, -1, -2 points, respectively. We add up scores from all five workers to collectively decide which set is better.

\subsection{The objectness map input for failure cases}
\bo{This section accompanies Sec. 4.3 in the main paper.}

\begin{figure}

\centering
\renewcommand{\tabcolsep}{0pt}
\includegraphics[width=1\columnwidth]{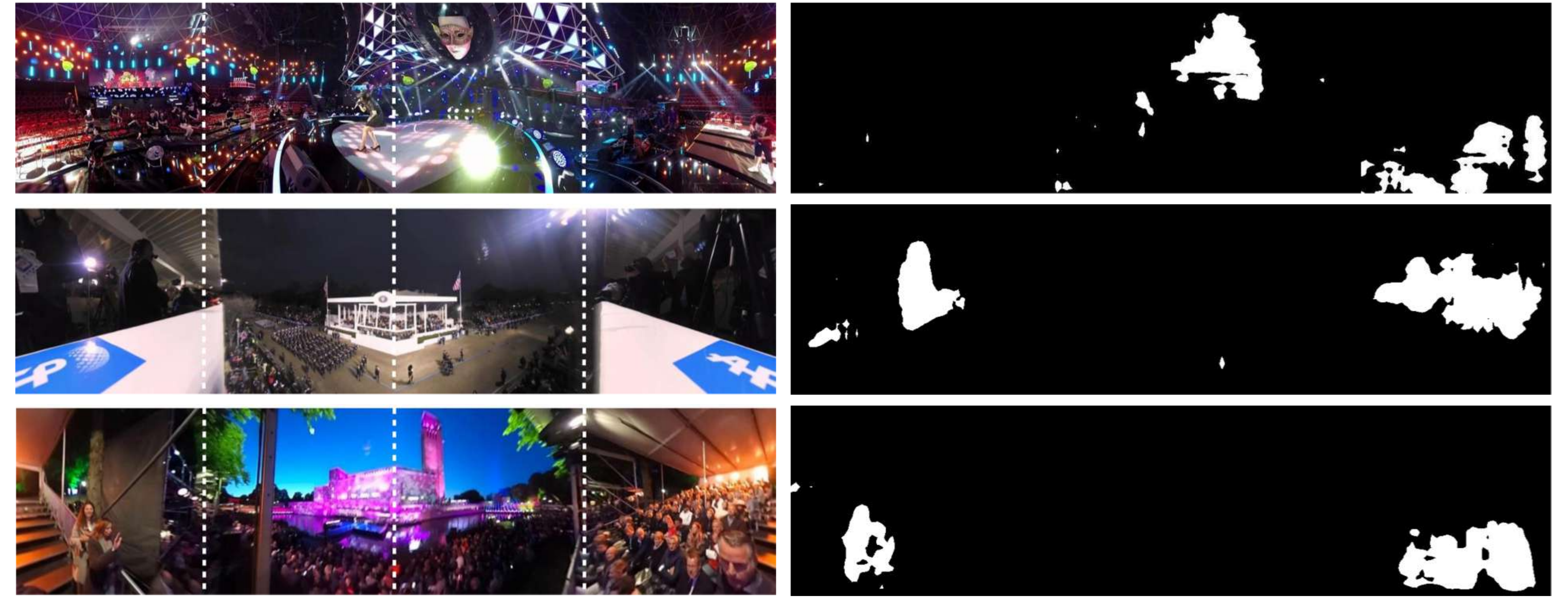}

\caption{\bo{Pixel objectness map (right) for failure cases of snap angle prediction. In the top row, the round stage is not found in the foreground, and so ends up distorted by our snap angle.}}
\label{fig:fail}
\end{figure}

\bo{Our method fails to preserve object integrity if pixel objectness fails to recognize foreground objects. Please see Fig.~\ref{fig:fail} for examples. The round stage is not found in the foreground, and so ends up distorted by our snap angle prediction method. In addition, the current solution cannot effectively handle the case when a foreground object is too large to fit in a single cube face. 
}

\clearpage

\subsection{Additional cubemap output examples}

This section accompanies Sec. 4.3 in the main paper.

\begin{figure*}[t]
\centering
\renewcommand{\tabcolsep}{0pt}
\includegraphics[width=1\columnwidth]{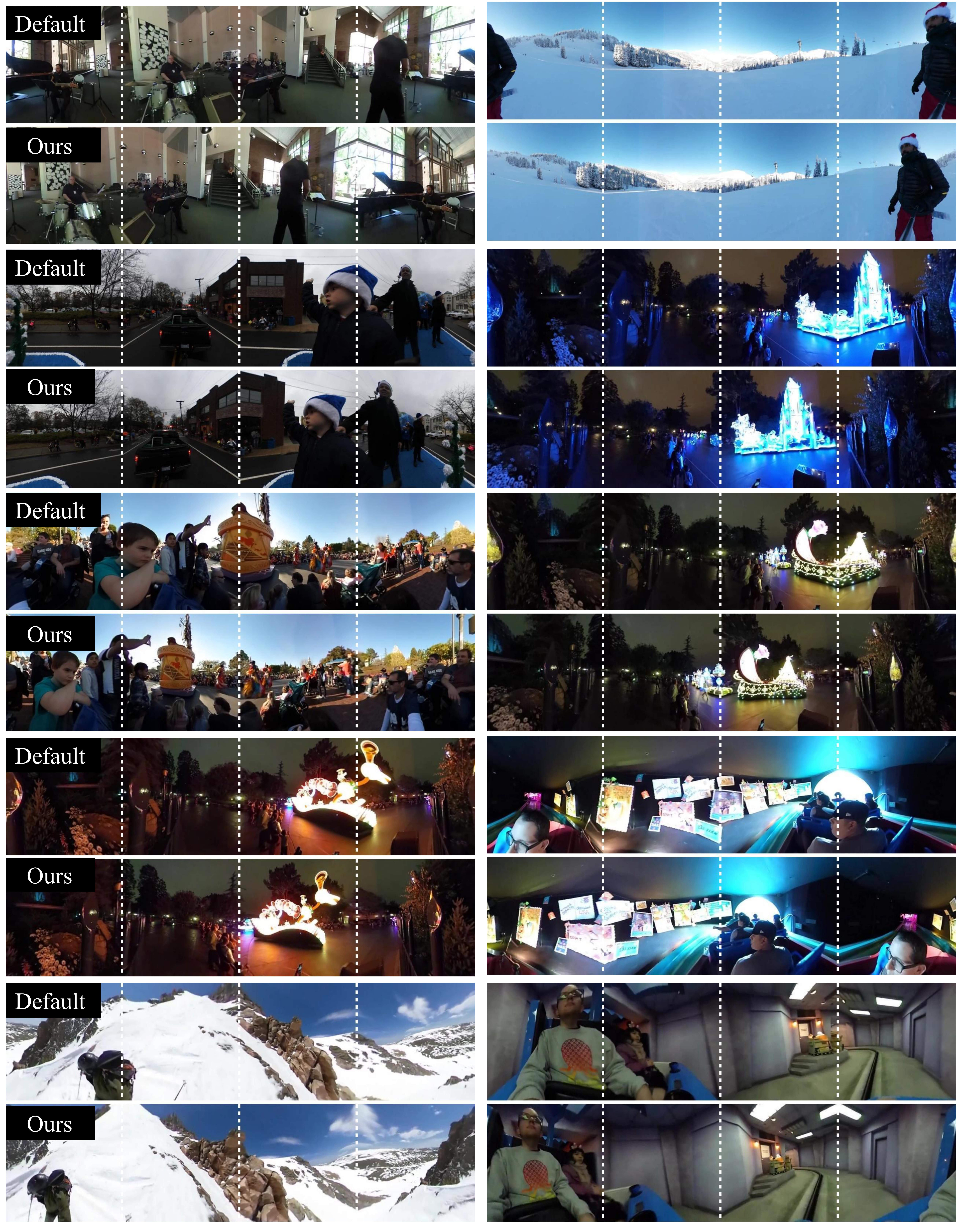}
\vspace{-10pt}
\caption{Qualitative examples of default \textsc{Canonical} cubemaps and our snap angle cubemaps.  Our method produces cubemaps that place important objects/persons in the same cube face to preserve the foreground integrity.}
\label{fig:supp_qual1}
\end{figure*}

Figure~\ref{fig:supp_qual1} presents additional cubemap examples.  
Our method produces cubemaps that place important objects/persons in the same cube face to preserve the foreground integrity. For example, in the top right of Figure~\ref{fig:supp_qual1}, our method places the person in a single cube face whereas the default cubemap splits the person onto two different cube faces.

\begin{figure*}[t]
\centering
\renewcommand{\tabcolsep}{0pt}
\includegraphics[width=1\columnwidth]{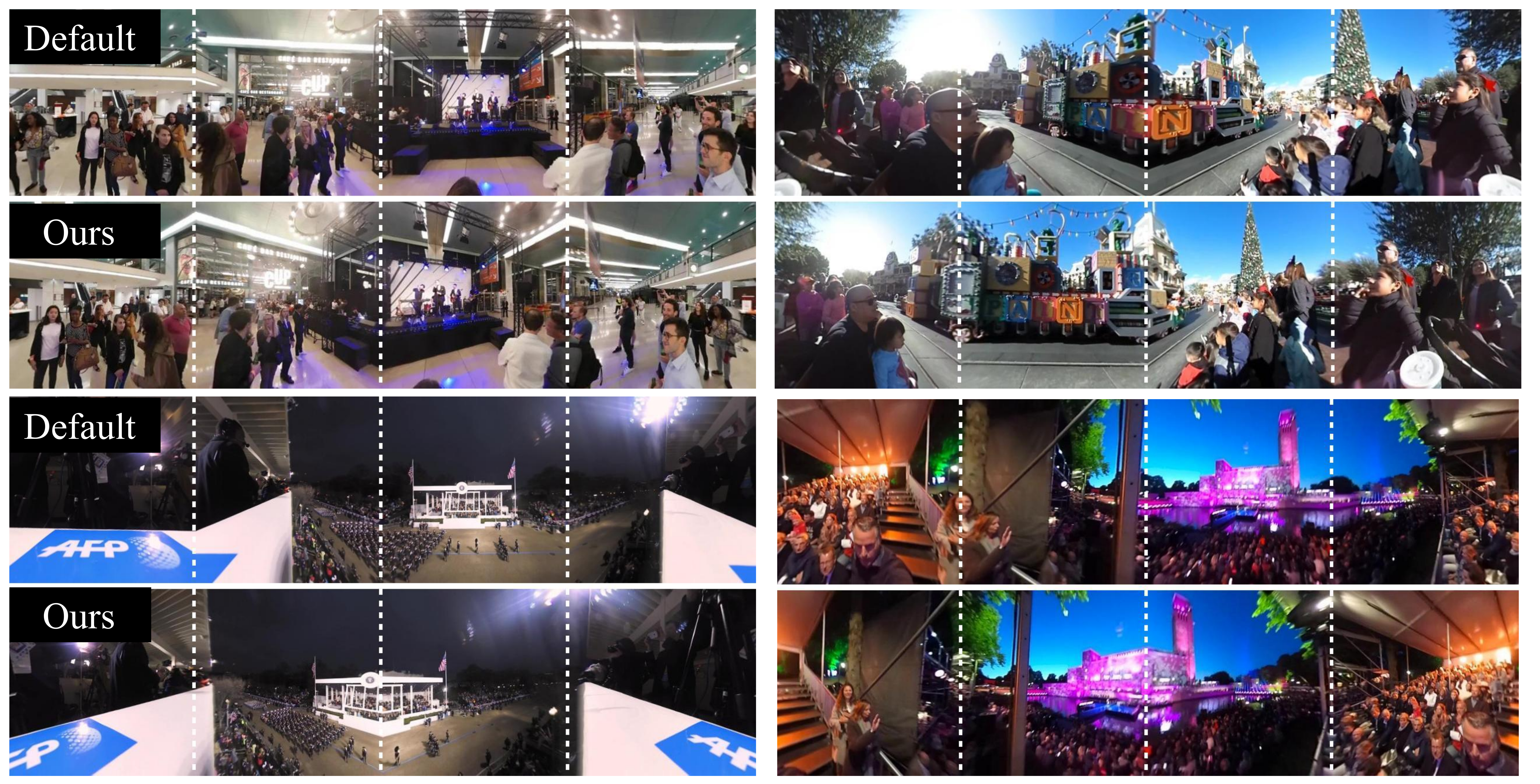}
\caption{Qualitative examples of default \textsc{Canonical} cubemaps and our snap angle cubemaps. We show failure cases here. In the top left, pixel objectness~\cite{jain2017pixel} does not recognize the stage as foreground, and therefore our method splits the stage onto two different cube faces, creating a distorted stage. In the top right, our method creates a distorted train by
splitting the train onto three different cube faces because pixel objectness does not recognize the train as foreground.}
\label{fig:supp_qual3}
\end{figure*}

Figure~\ref{fig:supp_qual3} shows failure cases.
A common reason for failure is that pixel objectness~\cite{jain2017pixel} does not recognize some important objects as foreground. For example, in the top right of Figure~\ref{fig:supp_qual3}, our method creates a distorted train by
splitting the train onto three different cube faces because pixel objectness does not recognize the train as foreground.

\subsection{Setup details for recognition experiment}

This section accompanies Sec. 4.4 in the main paper.

Recall our goal for the recognition experiment is to distinguish the four activity categories in our 360 dataset (Disney, Parade, etc.). We build a training dataset by querying Google Image Search engine for each activity category and then manually filtering out irrelevant images.  These are the same as the positive images in Sec.~\ref{sec:pano} above.

We use Resnet-101 architecture~\cite{he2016deep} as our classifier. The network is first pre-trained on Imagenet~\cite{russakovsky2015imagenet} and then we finetune
 it on our dataset with SGD and a mini-batch size of 32. The learning rate starts from 0.01 with a weight decay of 0.0001 and a momentum of 0.9. We train the network 
until convergence and select the best model based on a validation set.

\end{document}